\documentclass[10pt, journal]{IEEEtran}
\usepackage{graphicx}
\usepackage{multirow}
\usepackage{times}
\usepackage{adjustbox}
\usepackage{amsmath}
\usepackage{amssymb}
\usepackage{array}
\usepackage{color}
\usepackage{subcaption}
\usepackage{mathtools}
\usepackage{url}
\usepackage{bm}
\usepackage{breqn}
\usepackage{chronosys}

\usepackage{fancyhdr}
\fancypagestyle{firstpage}{%
  \rhead{IEEE Transactions on Circuits and Systems for Video Technology}
  \lfoot{\small{This paper is published by IEEE TCSVT (https://ieeexplore.ieee.org/document/9432822). Copyright © 2021 IEEE. Personal use of this material is permitted. However, permission to use this material for any other purposes must be obtained from the IEEE by sending an email to pubs-permissions@ieee.org. DOI of Final Paper: 10.1109/TCSVT.2021.3080920.}}
}

\begin{document}
\title{A Decade Survey of Content Based Image Retrieval using Deep Learning}

\author{
Shiv Ram Dubey, \IEEEmembership{Member,~IEEE}
\thanks{S.R. Dubey is with the Computer Vision Group, Indian Institute of Information Technology, Sri City, Chittoor, Andhra Pradesh-517646, India (e-mail: shivram1987@gmail.com, srdubey@iiits.in). }
}


\maketitle
\thispagestyle{firstpage}

\begin{abstract}
The content based image retrieval aims to find the similar images from a large scale dataset against a query image. Generally, the similarity between the representative features of the query image and dataset images is used to rank the images for retrieval. In early days, various hand designed feature descriptors have been investigated based on the visual cues such as color, texture, shape, etc. that represent the images. However, the deep learning has emerged as a dominating alternative of hand-designed feature engineering from a decade. It learns the features automatically from the data. This paper presents a comprehensive survey of deep learning based developments in the past decade for content based image retrieval. The categorization of existing state-of-the-art methods from different perspectives is also performed for greater understanding of the progress. The taxonomy used in this survey covers different supervision, different networks, different descriptor type and different retrieval type. A performance analysis is also performed using the state-of-the-art methods. The insights are also presented for the benefit of the researchers to observe the progress and to make the best choices. The survey presented in this paper will help in further research progress in image retrieval using deep learning. 
\end{abstract}


\section{Introduction}
\label{Introduction}
\IEEEPARstart{I}{mage} retrieval is a well studied problem of image matching where the similar images are retrieved from a database w.r.t. a given query image \cite{flickner1995query}, \cite{smeulders2000content}. Basically, the similarity between the query image and the database images is used to rank the database images in decreasing order of similarity \cite{muller2001performance}. Thus, the performance of any image retrieval method depends upon the similarity computation between images. Ideally, the similarity score computation method between two images should be discriminative, robust and efficient.

\subsection{Hand-crafted Descriptor based Image Retrieval}
In order to make the retrieval robust to geometric and photometric changes, the similarity between images is computed based on the content of images. Basically, the content of the images (i.e., the visual appearance) in terms of the color, texture, shape, gradient, etc. are represented in the form of a feature descriptor \cite{deselaers2008features}. The similarity between the feature vectors of the corresponding images is treated as the similarity between the images. Thus, the performance of any content based image retrieval (CBIR) method heavily depends upon the feature descriptor representation of the image. Any feature descriptor representation method is expected to have the discriminating ability, robustness and low dimensionality. 
Various feature descriptor representation methods have been investigated to compute the similarity between the two images for content based image retrieval. The feature descriptor representation utilizes the visual cues of the images selected manually based on the need \cite{sift}, \cite{lbp}, \cite{iold}, \cite{jacob2014local}, \cite{song2009biologically}, 
\cite{jegou2011aggregating}, \cite{ltrp}, \cite{lwp}, 
\cite{gong2012iterative}, 
\cite{lvp}, \cite{mdlbp}, \cite{chakraborty2016local}. 
These approaches are also termed as the hand-designed or hand-engineered feature description. Moreover, generally these methods are unsupervised as they do not need the data to design the feature representation method.
Various survey has been also conducted time to time to present the progress in content based image retrieval, including \cite{datta2008image} in 2008, \cite{mei2014multimedia} in 2014 and \cite{zhou2017recent} in 2017.
The hand-engineering feature for image retrieval was a very active research area. However, its performance was limited as the hand-engineered features are not able to represent the image characteristics in an accurate manner.

\begin{figure}[!t]
    \centering
    \includegraphics[trim=0 220 0 130, clip, width=\columnwidth]{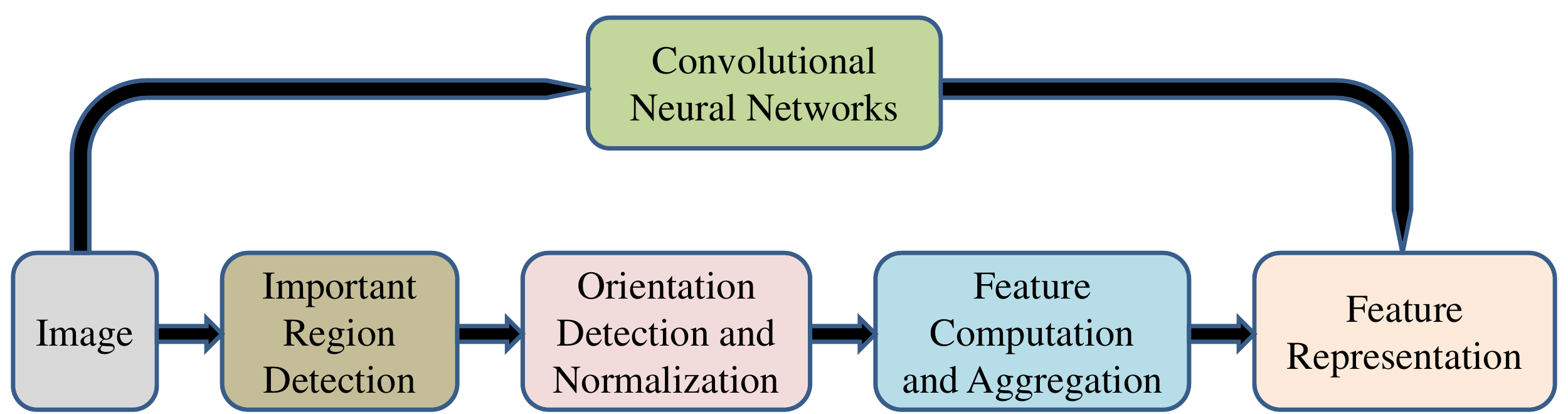}
    \caption{The pipeline of state-of-the-art feature representation is replaced by the CNN based feature representation. 
    }
    \label{fig:cnn_feature}
\end{figure}

\subsection{Distance Metric Learning based Image Retrieval}
The distance metric learning has been also used very extensively for feature vectors representation \cite{bellet2013survey}. It is also explored well for image retrieval \cite{wang2017survey}. Some notable deep metric learning based image retrieval approaches include
Contextual constraints distance metric learning \cite{hoi2006learning}, Kernel-based distance metric learning \cite{chang2007kernel}, \cite{liu2012supervised},
Visuality-preserving distance metric learning \cite{yang2008boosting},
Rank-based distance metric learning \cite{lee2008rank},
Semi-supervised distance metric learning \cite{hoi2010semi}, Hamming distance metric learning \cite{norouzi2012hamming}, \cite{song2016fast}, and Rank based metric learning \cite{song2015rank}, \cite{song2015top}.
Generally, the deep metric learning based approaches have shown the promising retrieval performance compared to hand-crafted approaches. However, most of the existing deep metric learning based methods rely on the linear distance functions which limits its discriminative ability and robustness to represent the non-linear data for image retrieval. Moreover, it is also not able to handle the multi-modal retrieval effectively.

\subsection{Deep Learning based Image Retrieval}
From a decade, a shift has been observed in feature representation from hand-engineering to learning-based after the emergence of deep learning \cite{sharif2014cnn}, \cite{jing2020self}. This transition is depicted in Fig. \ref{fig:cnn_feature} where the convoltional neural networks based feature learning replaces the state-of-the-art pipeline of traditional hand-engineered feature representation. The deep learning is a hierarchical feature representation technique to learn the abstract features from data which are important for that dataset and application \cite{lecun2015deep}. Based on the type of data to be processed, different architectures came into existence such as Artificial Neural Network (ANN)/ Multilayer Perceptron (MLP) for 1-D data 
\cite{dagli2012artificial}, \cite{amato2013artificial}, Convolutional Neural Networks (CNN) for image data \cite{alexnet}, 
\cite{resnet}, and Reurrent Neural Networks (RNN) for time-series data \cite{sundermeyer2012lstm}, \cite{chung2015gated}. 
A huge progress has been made in this decade to utilize the power of deep learning for content based image retrieval \cite{sharif2014cnn}, \cite{wan2014deep}, \cite{zheng2017sift}, \cite{rodrigues2020deep}, \cite{DHM_Survey}. Thus, this survey mainly focuses over the progress in state-of-the-art deep learning based models and features for content based image retrieval from its inception.
A taxonomy for the same is portrayed in Fig. \ref{fig:texonomy}. The major contributions of this survey can be outlined as follows:

\begin{figure}[!t]
    \centering
    \includegraphics[trim=0 20 85 0, clip, width=\columnwidth]{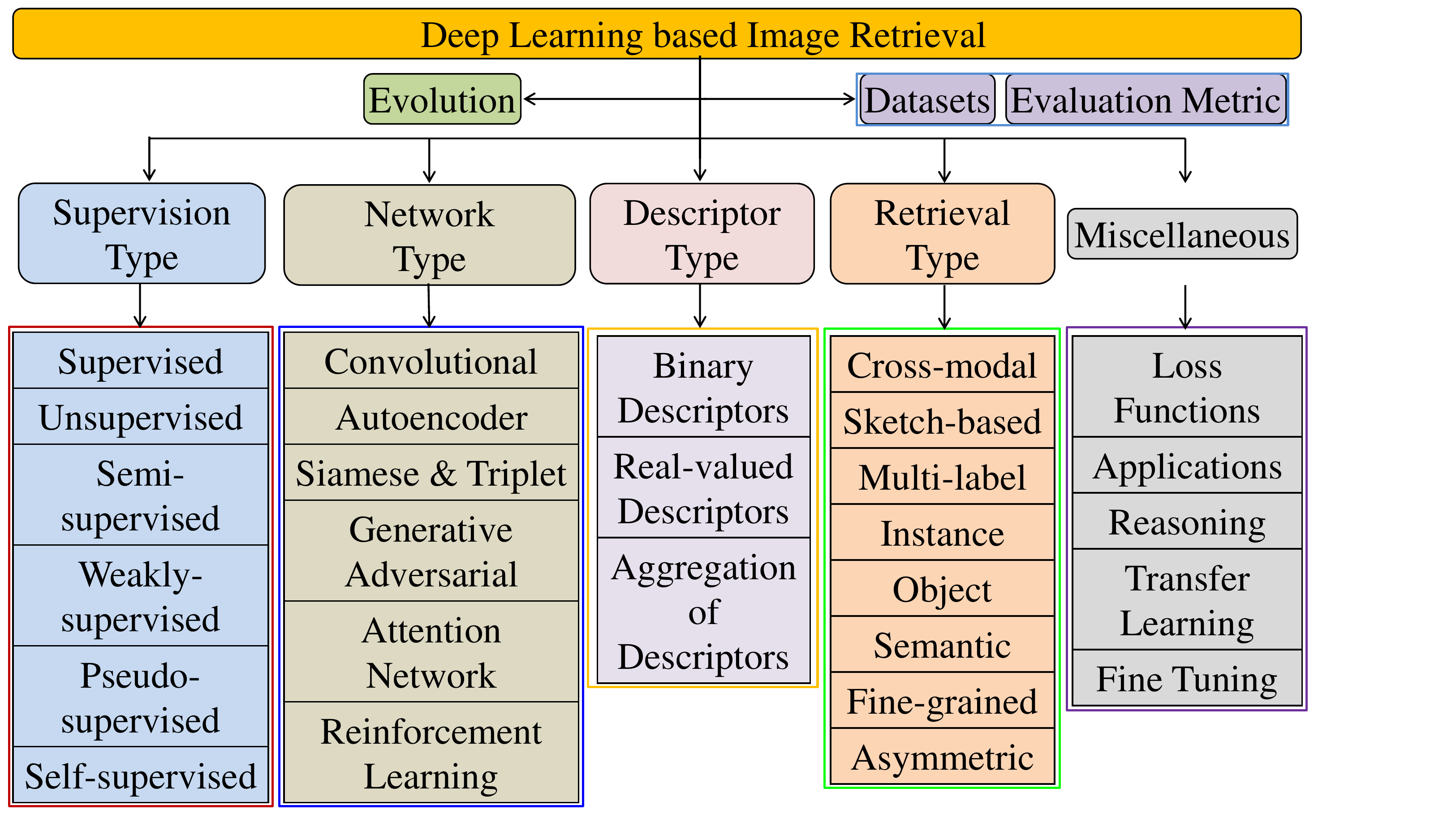}
    \caption{Taxonomy used in this survey to categorize the existing deep learning based image retrieval approaches.}
    \label{fig:texonomy}
\end{figure}

\begin{enumerate}
    \item This survey covers the deep learning based image retrieval approaches very comprehensively in terms of evolution of image retrieval using deep learning, different supervision type, network type, descriptor type, retrieval type and other aspects.
    \item In contrast to the recent reviews \cite{zheng2017sift}, \cite{wang2017survey}, \cite{rodrigues2020deep}, this survey specifically covers the progress in image retrieval using deep learning in 2011-2020 decade. An informative taxonomy is provided with wide coverage of existing deep learning based image retrieval approaches as compared to the recent survey \cite{DHM_Survey}.
    \item This survey enriches the reader with the state-of-the-art image retrieval using deep learning methods with analysis from various perspectives.
    \item This paper also presents the brief highlights and important discussions along with the comprehensive comparisons on benchmark datasets using the state-of-the-art deep learning based image retrieval approaches. 
\end{enumerate}

This survey is organized as follows: the background is presented in Section \ref{background} 
the evolution of deep learning based image retrieval is compiled in Section \ref{evolution}; the categorization of existing approaches based on the supervision type, network type, descriptor type, and retrieval type are discussed in Section \ref{supervision}, \ref{network}, \ref{descriptor}, and \ref{retrieval}, respectively; some other aspects are highlighted in Section \ref{misc}; the performance comparison of the popular methods is performed in Section \ref{performance_comparison}; conclusions and future directions are presented in Section \ref{conclusion}.

\section{Background} \label{background}
In this section the background is presented in terms of the commonly used evaluation metrics and benchmark datasets. 

\subsection{Retrieval Evaluation Measures}
In order to judge the performance of image retrieval approaches, precision, recall and f-score are the common evaluation metrics. 
The mean average precision ($mAP$) is very commonly used in the literature. The precision is defined as the percentage of correctly retrieved images out of the total number of retrieved images. The recall is another performance measure being used for image retrieval by computing the percentage of correctly retrieved images out of the total number of relevant images present in the dataset. The f-score is computed from the harmonic mean of precision and recall.

\begin{table}[!t]
    \centering
    \caption{The summary of large-scale datasets for deep learning based image retrieval.}
    \begin{adjustbox}{max width=\columnwidth}
    \begin{tabular}{|m{0.265\columnwidth}|m{0.042\columnwidth}|m{0.092\columnwidth}|m{0.103\columnwidth}|m{0.083\columnwidth}|m{0.331\columnwidth}|}
    \hline
        Dataset & Year & \#Classes & Training & Test & Image Type\\\hline
        CIFAR-10 \cite{cifar10} & 2009 & 10 & 50,000 & 10,000 & Object Category Images\\\hline
        NUS-WIDE \cite{nus-wide} & 2009 & 21 & 97,214 & 65,075 & Scene Images\\\hline
        MNIST \cite{mnist} & 1998 & 10 & 60,000 & 10,000 & Handwritten Digit Images\\\hline
        SVHN \cite{svhn} & 2011 & 10 & 73,257 & 26,032 & House Number Images\\\hline
        SUN397 \cite{sun} & 2010 & 397 & 100,754 & 8,000 & Scene Images\\\hline
        UT-ZAP50K \cite{yu2014fine} & 2014 & 4 & 42,025 & 8,000 & Shoes Images\\\hline
        Yahoo-1M \cite{lin2015deep} & 2015 & 116 & 1,011,723 & 112,363 & Clothing Images\\\hline
        ILSVRC2012 \cite{imagenet} & 2012 & 1,000 & $\sim$1.2 M & 50,000 & Object Category Images\\\hline
        MS COCO \cite{coco} & 2015 & 80 & 82,783 & 40,504 & Common Object Images\\\hline
        MIRFlicker-1M \cite{mirflicker} & 2010 & - & 1 M & - & Scene Images\\\hline
        Google Landmarks \cite{noh2017large} & 2017 & 15 K & $\sim$1 M & - & Landmark Images\\\hline
        Google Landmarks v2 \cite{weyand2020google} & 2020 & 200 K & 5 M & - & Landmark Images\\\hline
        Clickture
        \cite{hua2013clickage} & 2013 & 73.6 M & 40 M & - & Search Log\\
    \hline
    \end{tabular}
    \end{adjustbox}
    \label{tab:dataset}
\end{table}

\begin{figure*}[!t]
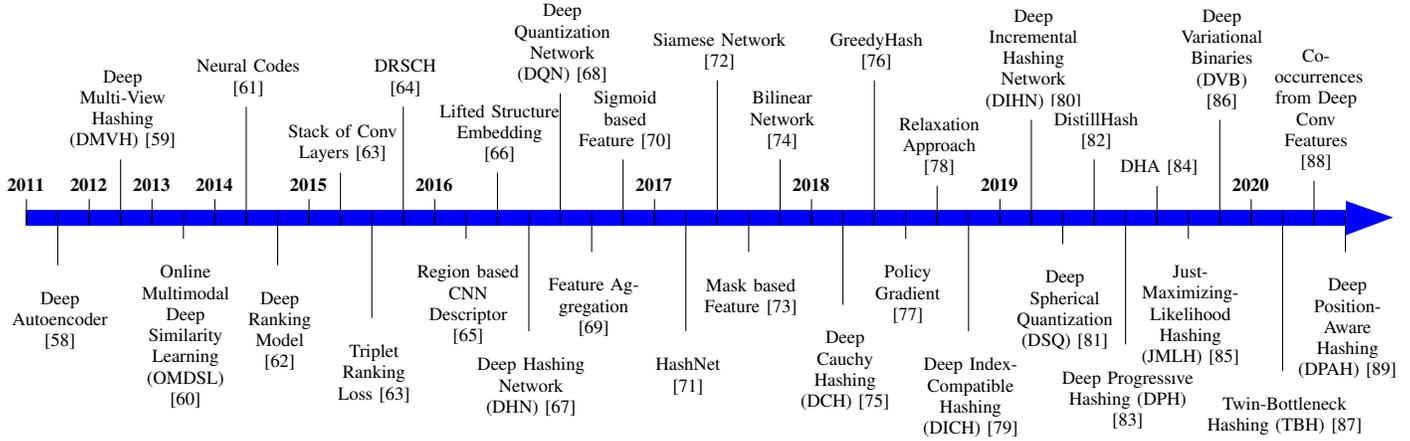

\centering
\vspace{-3.5cm}
\startchronology[startyear=1,stopyear=43, startdate=false, color=blue!100, height=0.2cm, stopdate=false, arrow=true, arrowwidth=0.6cm, arrowheight=0.45cm]
\setupchronoevent{textstyle=\scriptsize,datestyle=\scriptsize}
\chronoevent[markdepth=-15pt, year=false]{1}{\textbf{2011}}
\chronoevent[markdepth=15pt,year=false,textwidth=1.5cm]{2}{Deep Autoencoder \cite{krizhevsky2011using}}
\chronoevent[markdepth=-15pt, year=false]{3}{\textbf{2012}}
\chronoevent[markdepth=-30pt,year=false,textwidth=1.5cm]{4}{Deep Multi-View Hashing (DMVH) \cite{kang2012deep}}
\chronoevent[markdepth=-15pt, year=false]{5}{\textbf{2013}}
\chronoevent[markdepth=5pt,year=false,textwidth=1.5cm]{6}{Online Multimodal Deep Similarity Learning (OMDSL) \cite{wu2013online}}
\chronoevent[markdepth=-15pt, year=false]{7}{\textbf{2014}}
\chronoevent[markdepth=-50pt,year=false,textwidth=1.5cm]{8}{Neural Codes \cite{babenko2014neural}} 
\chronoevent[markdepth=15pt,year=false,textwidth=1cm]{9}{Deep Ranking Model \cite{wang2014learning}}
\chronoevent[markdepth=-15pt, year=false]{10}{\textbf{2015}}
\chronoevent[markdepth=-25pt,year=false,textwidth=1.5cm]{11}{Stack of Conv Layers \cite{lai2015simultaneous}}
\chronoevent[markdepth=35pt,year=false,textwidth=1cm]{12}{Triplet Ranking Loss \cite{lai2015simultaneous}}
\chronoevent[markdepth=-50pt,year=false,textwidth=1cm]{13}{DRSCH \cite{zhang2015bit}}
\chronoevent[markdepth=-15pt, year=false]{14}{\textbf{2016}}
\chronoevent[markdepth=5pt,year=false,textwidth=1.5cm]{15}{Region based CNN Descriptor \cite{gordo2016deep}}
\chronoevent[markdepth=-25pt,year=false,textwidth=1.6cm]{16}{Lifted Structure Embedding \cite{oh2016deep}}
\chronoevent[markdepth=40pt,year=false,textwidth=1.5cm]{17}{Deep Hashing Network (DHN) \cite{zhu2016deep}}
\chronoevent[markdepth=-55pt,year=false,textwidth=1.5cm]{18}{Deep Quantization Network (DQN) \cite{cao2016deep}}
\chronoevent[markdepth=10pt,year=false,textwidth=1.2cm]{19}{Feature Aggregation \cite{husain2016improving}}
\chronoevent[markdepth=-30pt,year=false,textwidth=1.3cm]{20}{Sigmoid based Feature \cite{zhong2016deep}}
\chronoevent[markdepth=-15pt, year=false]{21}{\textbf{2017}}
\chronoevent[markdepth=40pt,year=false,textwidth=1cm]{22}{HashNet \cite{cao2017hashnet}}
\chronoevent[markdepth=-60pt,year=false,textwidth=2cm]{23}{ Siamese Network \cite{gordo2017end}}
\chronoevent[markdepth=10pt,year=false,textwidth=1.3cm]{24}{Mask based Feature \cite{hoang2017selective}}
\chronoevent[markdepth=-30pt,year=false,textwidth=1.3cm]{25}{Bilinear Network \cite{alzu2017content}}
\chronoevent[markdepth=-15pt, year=false]{26}{\textbf{2018}}
\chronoevent[markdepth=30pt,year=false,textwidth=1.3cm]{27}{Deep Cauchy Hashing (DCH) \cite{cao2018deep}}
\chronoevent[markdepth=-60pt,year=false,textwidth=1.3cm]{28}{GreedyHash \cite{su2018greedy}}
\chronoevent[markdepth=5pt,year=false,textwidth=1.3cm]{29}{Policy Gradient \cite{yuan2018relaxation}}
\chronoevent[markdepth=-20pt,year=false,textwidth=1.3cm]{30}{Relaxation Approach \cite{chen2018deep}}
\chronoevent[markdepth=40pt,year=false,textwidth=1.3cm]{31}{Deep Index-Compatible Hashing (DICH) \cite{wu2018deep}}
\chronoevent[markdepth=-15pt, year=false]{32}{\textbf{2019}}
\chronoevent[markdepth=-45pt,year=false,textwidth=1.3cm]{33}{Deep Incremental Hashing Network (DIHN) \cite{wu2019deep}}
\chronoevent[markdepth=7pt,year=false,textwidth=1.3cm]{34}{Deep Spherical Quantization (DSQ) \cite{eghbali2019deep}}
\chronoevent[markdepth=-30pt,year=false,textwidth=1.3cm]{35}{DistillHash \cite{yang2019distillhash}}
\chronoevent[markdepth=45pt,year=false,textwidth=2cm]{36}{Deep Progressive Hashing (DPH) \cite{bai2019deep}}
\chronoevent[markdepth=-20pt,year=false,textwidth=1.3cm]{37}{DHA \cite{xu2019dha}}
\chronoevent[markdepth=5pt,year=false,textwidth=1.3cm]{38}{Just-Maximizing-Likelihood Hashing (JMLH) \cite{shen2019embarrassingly}}
\chronoevent[markdepth=-45pt,year=false,textwidth=1.1cm]{39}{Deep Variational Binaries (DVB) \cite{shen2019unsupervised}}
\chronoevent[markdepth=-15pt, year=false]{40}{\textbf{2020}}
\chronoevent[markdepth=55pt,year=false,textwidth=2.2cm]{41}{Twin-Bottleneck Hashing (TBH) \cite{shen2020auto}}
\chronoevent[markdepth=-22pt,year=false,textwidth=1.1cm]{42}{Co-occurrences from Deep Conv Features \cite{forcen2020co}}
\chronoevent[markdepth=10pt,year=false,textwidth=1.45cm]{43}{Deep Position-Aware Hashing (DPAH) \cite{wang2020deep}}
\stopchronology
\vspace{-3cm}
\caption{A chronological view of deep learning based image retrieval methods depicting its evolution from 2011 to 2020.}
\label{fig:evolution}
\end{figure*}

\subsection{Datasets}
With the inception of deep learning models, various large-scale datasets have been created to facilitate the research in image recognition and retrieval. The details of large-scale datasets are summarized in Table \ref{tab:dataset}. Datasets having various types of images are available to test the deep learning based approaches such as object category datasets \cite{cifar10}, \cite{imagenet}, \cite{coco}, scene datasets \cite{nus-wide}, \cite{sun}, \cite{flicker}, digit datasets \cite{mnist}, \cite{svhn}, apparel datasets \cite{yu2014fine}, \cite{lin2015deep}, landmark datasets \cite{noh2017large}, \cite{weyand2020google}, etc. 
The CIFAR-10 dataset is very widely used object category datset \cite{cifar10}. The ImageNet (ILSVRC2012), a large-scale dataset, is also an object category dataset with more than a million number of images \cite{imagenet}. The MS COCO dataset \cite{coco} created for common object detection is also utilized for image retrieval purpose.
Among scene image datasets commonly used for retrieval purpose, the NUS-WIDE dataset is from National University of Singapore \cite{nus-wide}; the Sun397 is a scene understanding dataset from 397 categories with more than one lakh images \cite{sun}, \cite{sun1}; and the MIRFlicker-1M \cite{flicker} dataset consists of a million images downloaded from the social photography site Flickr. The MNIST dataset is one of the old and large-scale digit image datasets \cite{mnist} consisting of optical characters. The SVHN is another digit dataset \cite{svhn} from the street view house number images which is more complex than MNIST dataset. The shoes apparel dataset, namely UT-ZAP50K \cite{yu2014fine}, consists of roughly 50K images. The Yahoo-1M is another apparel large-scale dataset used in \cite{lin2015deep} for image retrieval. The Google landmarks dataset is having around a million landmark images \cite{noh2017large}. The extended version of Google landmarks (i.e., v2) \cite{weyand2020google} contains around 5 million landmark images. There are more datasets used for retrieval in the literature, such as Corel, Oxford, Paris, etc., however, these are not the large-scale datasets. The CIFAR-10, MNIST, SVHN and ImageNet are the widely used datasets in majority of the research. 
Clickture is a common dataset for search log based on the queries of users \cite{hua2013clickage}. The click property has been utilized for different applications, such as cross-view learning for image search \cite{pan2014click}, distance metric learning for image ranking \cite{yu2016deep} and deep structure-preserving embeddings with visual attention \cite{li2019learning}.

Note that only CIFAR-10 and MNIST datasets contain the same number of samples in each category. Other datasets are created generally in unconstrained environment with huge number of samples, thus the classes are not well balanced. The choice of dataset can be dependent upon the scenario where image retrieval models need to be used, such as object category and scene datasets for unconstrained environment, apparel datasets for e-commerce applications, and landmark datasets for driving applications.

\section{Evolution of Deep Learning for Content Based Image Retrieval (CBIR)} \label{evolution}
The deep learning based generation of descriptors or hash codes is the recent trends large-scale content based image retrieval, due to its computational efficiency and retrieval quality \cite{wang2017survey}. 
In this section, a journey of deep learning models for image retrieval from 2011 to 2020 is presented as a chronological overview in Fig. \ref{fig:evolution}.

\subsubsection{2011-2013}
Among the initial attempts, in 2011, Krizhevsky and Hinton have used a deep autoencoder to map the images to short binary codes for content based image retrieval (CBIR) \cite{krizhevsky2011using}. Kang et al. (2012) have proposed a deep multi-view hashing to generate the code for CBIR from multiple views of data by modeling the layers with view-specific and shared hidden nodes \cite{kang2012deep}. In 2013, Wu et al. have considered the multiple pretrained stacked denoising autoencoders over low features of the images \cite{wu2013online}. They also fine tune the multiple deep networks on the output of the pretrained autoencoders.

\subsubsection{2014}
\begin{figure}[!t]
    \centering
    \includegraphics[trim=10 270 80 10, clip, width=0.8\columnwidth]{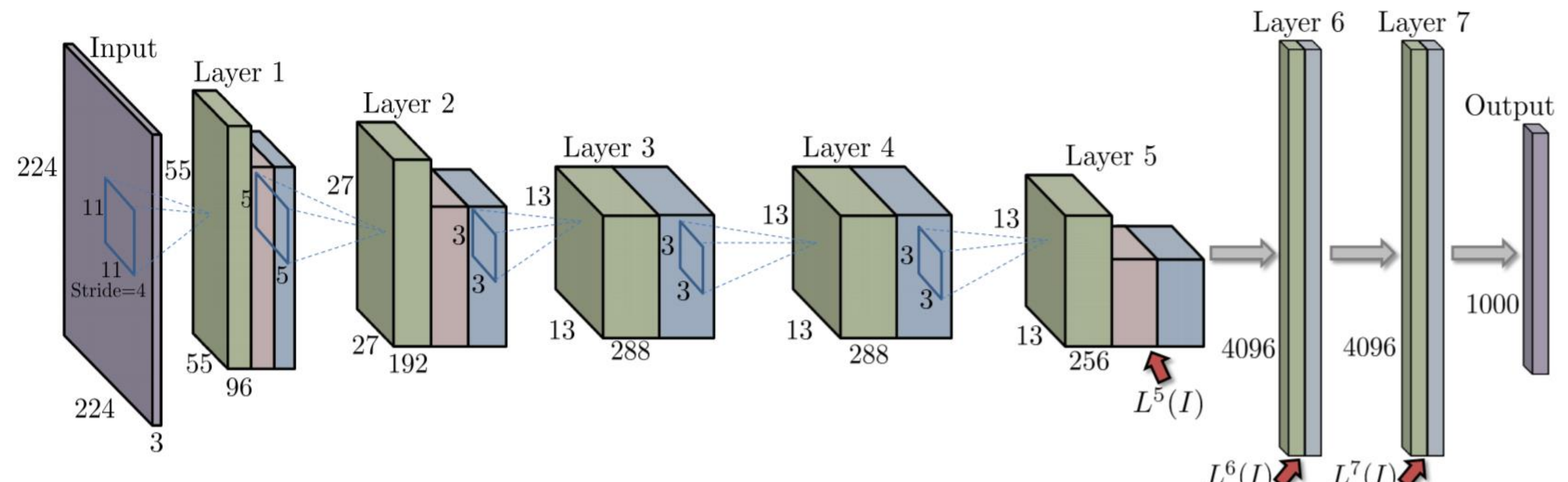}
    \caption{The illustration of the neural code generation from a convolutional neural network (CNN) \cite{babenko2014neural}. 
    }
    \label{fig:neural_code}
\end{figure}

In an outstanding work, the activations of the top layers of a large convolutional neural network (CNN) are utilized as the descriptors (neural codes) for image retrieval \cite{babenko2014neural} as depicted in Fig. \ref{fig:neural_code}. A very promising performance has been recorded using the neural codes for image retrieval even if the model is trained on un-related data. 
The neural code is compressed using principal component analysis (PCA) to generate the compact descriptor.
In 2014, deep ranking model is investigated by learning the similarity metric directly from images \cite{wang2014learning}. Basically, the triplets are employed to capture the inter-class and intra-class image differences.

\subsubsection{2015}
In 2015, a deep architecture is developed which consists of a stack of convolution layers to produce the intermediate image features \cite{lai2015simultaneous} which are used to generate the hash bits.
The triplet ranking loss is also utilized to incorporate the inter-class and intra-class differences in \cite{lai2015simultaneous} for image retrieval. Zhang et al. (2015) have developed a deep regularized similarity comparison hashing (DRSCH) by training a deep CNN model 
to simultaneously optimize the discriminative image features and hash functions \cite{zhang2015bit}.

\subsubsection{2016}
In 2016, Gordo et al. have pooled the relevant regions to form the descriptor with the help of a region proposal network to prioritize the important object regions \cite{gordo2016deep}.
Song et al. (2016) have computed the lifted structure loss between the CNN and the original features \cite{oh2016deep}.
Supervised deep hashing network (DHN) learns the important image representation by controlling the quantization error \cite{zhu2016deep}. At the same time, Cao et al. have introduced a deep quantization network (DQN) which is very similar to the DHN model \cite{cao2016deep}. 
The CNN based features are aggregated in \cite{husain2016improving} with the help of rank-aware multi-assignment and direction based combination. A sigmoid layer is added before the loss layer of a CNN to learn the binary code for CBIR \cite{zhong2016deep}.

\subsubsection{2017}
In 2017, Cao et al. have proposed HashNet deep architecture to generate the hash code by a continuation method \cite{cao2017hashnet}. It learns the non-smooth binary activations using the continuation method to generate the binary hash codes from imbalanced similarity data. Gordo et al. (2017) have shown that the noisy training data, inappropriate deep architecture and suboptimal training procedure are the main hurdle to utilize the deep learning for image retrieval \cite{gordo2017end}. 
Different masking schemes are used in \cite{hoang2017selective} to select the prominent CNN features for image retrieval. A bilinear network with two parallel CNNs is also used as a feature extractors \cite{alzu2017content}.

\subsubsection{2018}
In 2018, Cao et al. have investigated a deep cauchy hashing (DCH) model for binary hash code with the help of a pairwise cross-entropy loss based on Cauchy distribution \cite{cao2018deep}. Su et al. have employed the greedy hash by transmitting the gradient as intact during the backpropagation for hash coding layer which uses the sign function in forward propagation \cite{su2018greedy}. 
Different approaches such as policy gradient \cite{yuan2018relaxation} and series expansion \cite{chen2018deep} are also utilized to train the models.
Deep index-compatible hashing (DICH) method \cite{wu2018deep} is investigated by minimizing the number of similar bits between the binary codes of inter-class images.

\subsubsection{2019}
In 2019, a deep incremental hashing network (DIHN) is proposed in \cite{wu2019deep} to directly learn the hash codes corresponding to the new class coming images, while retaining the hash codes of existing class images. A supervised quantization based points representation on a unit hypersphere is used in deep spherical quantization (DSQ) model \cite{eghbali2019deep}. DistillHash \cite{yang2019distillhash} distills data pairs and learns deep hash functions from the distilled data set by employing the Bayesian learning framework.
A deep progressive hashing (DPH) model is developed to generate a sequence of binary codes by utilizing the progressively expanded salient regions \cite{bai2019deep}. 
Adaptive loss function based deep hashing \cite{xu2019dha}, just-maximizing-likelihood hashing (JMLH) \cite{shen2019embarrassingly} and deep variational binaries (DVB) \cite{shen2019unsupervised} are other approaches discovered in 2019.

\subsubsection{2020}
Recently, in 2020, Shen et al. have come up with a twin-bottleneck hashing (TBH) model between encoder and decoder networks \cite{shen2020auto}. They have employed the binary and continuous  bottlenecks as the latent variables in a collaborative manner. 
Forcen et al. (2020) have utilized the last convolution layer of CNN representation by modeling the co-occurrences from deep convolutional features \cite{forcen2020co}.
A deep position-aware hashing (DPAH) model is proposed in 2020 \cite{wang2020deep} which constraints the distance between data samples and class centers.

Most of the methods developed between 2011 and 2015 use the features learnt by the autoencoders and convolutional neural networks. However, these methods face the issues in terms of the less discriminative ability as the models are generally trained for the classification problem and the information loss due to the quantization of features. Image retrieval using deep learning has witnessed a huge growth in between 2016 and 2020. As the image retrieval application needs feature learning for matching, different types of networks have been utilized to do so. The recent methods have designed the several objective functions which lead to the high inter class separation and high intra class condensation in feature space. Moreover, the development in different network architectures has also led to the growth in the image retrieval area. The key issue being addressed by deep learning methods is to learn very discriminative, robust and compact features for image retrieval.

\section{Different Supervision Categorization} \label{supervision}
This section covers the image retrieval methods in terms of the different supervision types. Basically, supervised, unsupervised, semi-supervised, weakly-supervised, pseudo-supervised and self-supervised approaches are included. 

\subsection{Supervised Approaches}
The supervised deep learning models are used by researchers very heavily to learn the class specific and discriminative features for image retrieval. In 2014, Xia et al. have used a CNN to learn the representation of images which is used to generate a hash code and class labels \cite{xia2014supervised}. 
The promising performance is reported over MNIST, CIFAR-10 and NUS-WIDE datasets. 
Shen et al. (2015) \cite{shen2015supervised} have proposed the supervised discrete hashing (SDH) based generation of image description with the help of the discrete cyclic coordinate descent for retrieval.
Liu et al. (2016) have done the revolutionary work by introducing a deep supervised hashing (DSH) method to learn the binary codes from the similar/dissimilar pairs of images \cite{liu2016deep}. 
A similar work is also presented in deep pairwise-supervised hashing (DPSH) method for image retrieval \cite{li2016feature}. 
The pair-wise labels are extended to the triplet labels (i.e., query, positive and negative images) to train a shared deep CNN model for feature learning \cite{wang2016deep}. 
An independent layer-wise local updates are performed in \cite{zhang2016efficient} to efficiently train a very deep supervised hashing (VDSH) model.

In 2017, Li et al. have used the classification information and the pairwise label information in a single framework for the learning of the deep supervised discrete hashing (DSDH) codes \cite{li2017deep}. 
The supervised semantics-preserving deep hashing (SSDH) model integrates the retrieval and classification characteristics in feature learning
\cite{yang2017supervised}. 
The scalable image search is performed in \cite{lu2017deep} by introducing the following three characteristics: 1) minimizing the loss between the real-valued code and equivalent converted binary code, 2) ensuring the even distribution among each bit in the binary codes, and 3) decreasing the redundancy of a bit in the binary code.

The supervised training has been also the choice in asymmetric hashing \cite{jiang2018asymmetric}. A deep product quantization (DPQ) model is followed in supervised learning mode for image search and retrieval \cite{klein2019end}. The supervised deep feature embedding is also used with the hand crafted features \cite{kan2019supervised}. A very recently, a multi-Level hashing of deep features is performed by Ng et al. (2020) \cite{ng2020multi}. An angular hashing loss function is used to train the network in the supervised fashion \cite{zhou2019angular}. A supervised hashing is also used for the multi-deep ranking \cite{li2019weighted} to improve the retrieval efficiency. Other supervised approaches are deep binary hash codes \cite{lin2015deep}, deep hashing network \cite{zhu2016deep}, deep spherical quantization \cite{eghbali2019deep}, and adaptive loss based supervised deep learning to hash \cite{xu2019dha}.

\subsection{Unsupervised Approaches}
Though the supervised models have shown promising performance for retrieval, it is difficult to get the labelled large-scale data always. Thus, several unsupervised models have been also investigated which do not require the class labels. The unsupervised models generally enforce the constraints on hash code and/or generated output to learn the features.

Erin et al. (2015) \cite{erin2015deep} have used the deep networks in an unsupervised manner to learn the hash code with the help of the constraints like quantization loss, balanced bits and independent bits.
Huang et al. (2016) \cite{huang2016unsupervised} have utilized the CNN coupled with unsupervised discriminative clustering.
In an outstanding work, DeepBit utilizes the constraints like minimal quantization loss, evenly distributed codes and uncorrelated bits for unsupervised image retrieval \cite{lin2016learning}, \cite{lin2018unsupervised}.
In order to improve the robustness of DeepBit, a rotation data augmentation based fine tuning is also performed. 
However, the DeepBit model suffers with the severe quantization loss due to the rigid binarization of data using sign function without considering its distribution property.
Deep binary descriptor with multiquantization (DBD-MQ) \cite{duan2017learning} tackles the quantization problem of DeepBit by jointly learning the parameters and the binarization functions using a K-AutoEncoders (KAEs).

It is observed in \cite{radenovic2016cnn} that unsupervised CNN can learn more distinctive features if fine tuned with hard positive and hard negative examples. 
The patch representation using a patch convolutional kernel network is also adapted for patch retrieval  \cite{paulin2017convolutional}. 
An anchor image, a rotated image and a random image based triplets are used in unsupervised triplet hashing (UTH) to learn the binary codes for image retrieval \cite{huang2017unsupervised}. The UTH objective function uses the combination of discriminative loss, quantization loss and entropy loss.
An unsupervised similarity-adaptive deep hashing (SADH) is proposed in \cite{shen2018unsupervised} by updating a similarity graph and optimizing the binary codes. 
Xu et al. (2018) \cite{xu2018unsupervised} have proposed a semantic-aware part weighted aggregation using part-based detectors for CBIR systems. 
Unsupervised generative adversarial networks \cite{ghasedi2018unsupervised}, \cite{song2018binary}, \cite{deng2019unsupervised} are also investigated for image retrieval. The distill data pairs \cite{yang2019distillhash} and deep variational networks \cite{shen2019unsupervised} are also used for unsupervised image retrieval. The pseudo triplets based unsupervised deep triplet hashing (UDTH) technique \cite{gu2019unsupervised} is introduced for scalable image retrieval. Very recently unsupervised deep transfer learning has been exploited by Liu et al. (2020) \cite{liu2020similarity} for retrieval in remote sensing images.

Though the unsupervised models do not need labelled data, its performance is generally lower than the supervised approaches. Thus, researchers have explored the models between supervised and unsupervised, such as semi-supervised, weakly-supervised, pseudo-supervised and self-supervised.

\begin{figure*}[!t]
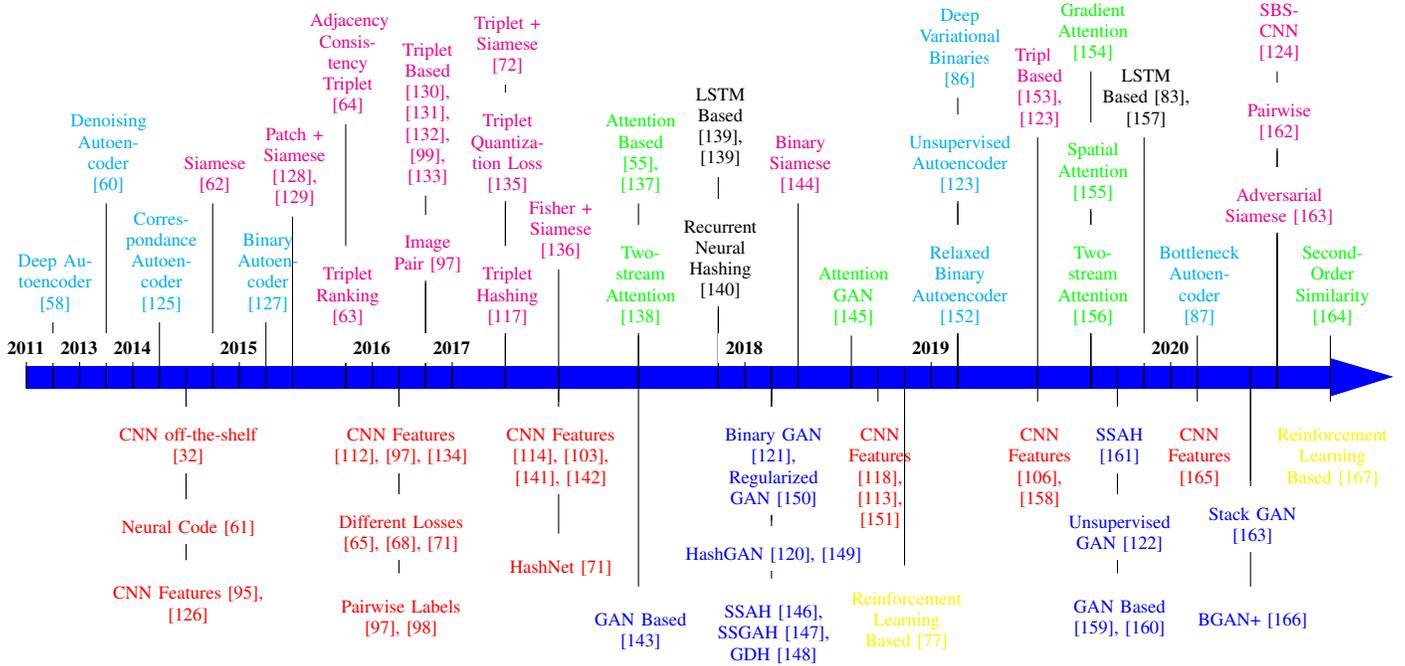

\centering
\vspace{-5.4cm}
\startchronology[startyear=1,stopyear=50, startdate=false, color=blue!100, stopdate=false, arrow=true, arrowwidth=0.8cm, arrowheight=0.5cm]
\setupchronoevent{textstyle=\scriptsize,datestyle=\scriptsize}
\chronoevent[markdepth=-15pt, year=false]{1}{\textbf{2011}}
\chronoevent[markdepth=-30pt,year=false,textwidth=1cm]{2}{\textcolor{cyan}{Deep Autoencoder \cite{krizhevsky2011using}}}
\chronoevent[markdepth=-15pt, year=false]{3}{\textbf{2013}}
\chronoevent[markdepth=-75pt,year=false,textwidth=1cm]{4}{\textcolor{cyan}{Denoising Autoencoder \cite{wu2013online}}}

\chronoevent[markdepth=-15pt, year=false]{5}{\textbf{2014}}
\chronoevent[markdepth=-30pt,year=false,textwidth=1cm]{6}{\textcolor{cyan}{Corres-pondance Autoencoder \cite{feng2014cross}}}
\chronoevent[markdepth=-75pt,year=false,textwidth=1cm]{8}{\textcolor{magenta}{Siamese \cite{wang2014learning}}}

\chronoevent[markdepth=65pt,year=false,textwidth=2cm]{7}{\textcolor{red}{CNN Features \cite{xia2014supervised}, \cite{sun2014search}}}
\chronoevent[markdepth=40pt,year=false,textwidth=2cm]{7}{\textcolor{red}{Neural Code \cite{babenko2014neural}}}
\chronoevent[markdepth=5pt,year=false,textwidth=2cm]{7}{\textcolor{red}{CNN off-the-shelf \cite{sharif2014cnn}}}

\chronoevent[markdepth=-15pt, year=false]{9}{\textbf{2015}}
\chronoevent[markdepth=-30pt,year=false,textwidth=1cm]{10}{\textcolor{cyan}{Binary Autoencoder \cite{carreira2015hashing}}}
\chronoevent[markdepth=-70pt,year=false,textwidth=1cm]{11}{\textcolor{magenta}{Patch + Siamese \cite{zagoruyko2015learning}, \cite{han2015matchnet}}}
\chronoevent[markdepth=-105pt,year=false,textwidth=1cm]{13}{\textcolor{magenta}{Adjacency Consistency Triplet \cite{zhang2015bit}}}
\chronoevent[markdepth=-25pt,year=false,textwidth=1cm]{13}{\textcolor{magenta}{Triplet Ranking \cite{lai2015simultaneous}}}

\chronoevent[markdepth=-15pt, year=false]{14}{\textbf{2016}}
\chronoevent[markdepth=-78pt,year=false,textwidth=1cm]{16}{\textcolor{magenta}{Triplet Based \cite{zhuang2016fast}, \cite{yao2016deep}, \cite{kumar2016learning}, \cite{wang2016deep}, \cite{lin2016tiny}}}
\chronoevent[markdepth=-45pt,year=false,textwidth=1cm]{16}{\textcolor{magenta}{Image Pair \cite{liu2016deep}}}

\chronoevent[markdepth=70pt,year=false,textwidth=2cm]{15}{\textcolor{red}{Pairwise Labels \cite{liu2016deep}, \cite{li2016feature}}}
\chronoevent[markdepth=38pt,year=false,textwidth=2cm]{15}{\textcolor{red}{Different Losses \cite{gordo2016deep}, \cite{cao2016deep}, \cite{cao2017hashnet}}}
\chronoevent[markdepth=5pt,year=false,textwidth=2cm]{15}{\textcolor{red}{CNN Features \cite{lin2016learning}, \cite{liu2016deep}, \cite{wei2016cross}}}

\chronoevent[markdepth=-15pt, year=false]{17}{\textbf{2017}}
\chronoevent[markdepth=-120pt,year=false,textwidth=1cm]{19}{\textcolor{magenta}{Triplet + Siamese \cite{gordo2017end}}}
\chronoevent[markdepth=-75pt,year=false,textwidth=1cm]{19}{\textcolor{magenta}{Triplet Quantization Loss \cite{zhou2017deep}}}
\chronoevent[markdepth=-25pt,year=false,textwidth=1cm]{19}{\textcolor{magenta}{Triplet Hashing \cite{huang2017unsupervised}}}
\chronoevent[markdepth=-50pt,year=false,textwidth=1cm]{21}{\textcolor{magenta}{Fisher + Siamese \cite{ong2017siamese}}}
\chronoevent[markdepth=-75pt,year=false,textwidth=1cm]{24}{\textcolor{green}{Attention Based \cite{noh2017large}, \cite{song2017deep}}}
\chronoevent[markdepth=-25pt,year=false,textwidth=1cm]{24}{\textcolor{green}{Two-stream Attention \cite{yang2017two}}}
\chronoevent[markdepth=-85pt,year=false,textwidth=1cm]{27}{\textcolor{black}{LSTM Based \cite{shen2017deep}, \cite{shen2017deep}}}
\chronoevent[markdepth=-35pt,year=false,textwidth=1cm]{27}{\textcolor{black}{Recurrent Neural Hashing \cite{lu2017hierarchical}}}

\chronoevent[markdepth=55pt,year=false,textwidth=1.5cm]{21}{\textcolor{red}{HashNet \cite{cao2017hashnet}}}
\chronoevent[markdepth=5pt,year=false,textwidth=1.5cm]{21}{\textcolor{red}{CNN Features \cite{duan2017learning}, \cite{lu2017deep}, \cite{zhang2017ssdh}, \cite{yang2017pairwise}}}
\chronoevent[markdepth=75pt,year=false,textwidth=1.5cm]{24}{\textcolor{blue}{GAN Based \cite{wang2017adversarial}}}

\chronoevent[markdepth=-15pt, year=false]{28}{\textbf{2018}}
\chronoevent[markdepth=-75pt,year=false,textwidth=1cm]{30}{\textcolor{magenta}{Binary Siamese \cite{jose2017binary}}}
\chronoevent[markdepth=-25pt,year=false,textwidth=1cm]{32}{\textcolor{green}{Attention GAN \cite{zhang2018attention}}}

\chronoevent[markdepth=72pt,year=false,textwidth=2cm]{29}{\textcolor{blue}{SSAH \cite{li2018self}, SSGAH \cite{wang2018semi}, GDH \cite{zhang2018generative}}}
\chronoevent[markdepth=50pt,year=false,textwidth=2.5cm]{29}{\textcolor{blue}{HashGAN \cite{ghasedi2018unsupervised}, \cite{cao2018hashgan}}}
\chronoevent[markdepth=5pt,year=false,textwidth=1.5cm]{29}{\textcolor{blue}{Binary GAN \cite{song2018binary}, Regularized GAN \cite{zieba2018bingan}}}
\chronoevent[markdepth=5pt,year=false,textwidth=1cm]{33}{\textcolor{red}{CNN Features \cite{shen2018unsupervised}, \cite{lin2018unsupervised}, \cite{yu2018multi}}}
\chronoevent[markdepth=67pt,year=false,textwidth=1.5cm]{34}{\textcolor{yellow}{Reinforcement Learning Based \cite{yuan2018relaxation}}}

\chronoevent[markdepth=-15pt, year=false]{35}{\textbf{2019}}
\chronoevent[markdepth=-115pt,year=false,textwidth=1.2cm]{36}{\textcolor{cyan}{Deep Variational Binaries \cite{shen2019unsupervised}}}
\chronoevent[markdepth=-75pt,year=false,textwidth=1.5cm]{36}{\textcolor{cyan}{Unsupervised Autoencoder \cite{gu2019unsupervised}}}
\chronoevent[markdepth=-25pt,year=false,textwidth=1.5cm]{36}{\textcolor{cyan}{Relaxed Binary Autoencoder \cite{do2019simultaneous}}}
\chronoevent[markdepth=-100pt,year=false,textwidth=1.2cm]{39}{\textcolor{magenta}{Triplet Based \cite{dey2019doodle}, \cite{gu2019unsupervised}}}
\chronoevent[markdepth=-125pt,year=false,textwidth=1cm]{41}{\textcolor{green}{Gradient Attention \cite{huang2019accelerate}}}
\chronoevent[markdepth=-72pt,year=false,textwidth=1cm]{41}{\textcolor{green}{Spatial Attention \cite{ge2019deep}}}
\chronoevent[markdepth=-25pt,year=false,textwidth=1cm]{41}{\textcolor{green}{Two-stream Attention \cite{wei2019saliency}}}
\chronoevent[markdepth=-100pt,year=false,textwidth=1.2cm]{43}{\textcolor{black}{LSTM Based \cite{bai2019deep}, \cite{chen2019beyond}}}

\chronoevent[markdepth=5pt,year=false,textwidth=1cm]{39}{\textcolor{red}{CNN Features \cite{kan2019supervised}, \cite{wang2019enhancing}}}
\chronoevent[markdepth=70pt,year=false,textwidth=1.5cm]{42}{\textcolor{blue}{GAN Based \cite{kumar2019generative}, \cite{gu2019adversary}}}
\chronoevent[markdepth=38pt,year=false,textwidth=1.5cm]{42}{\textcolor{blue}{Unsupervised GAN \cite{deng2019unsupervised}}}
\chronoevent[markdepth=5pt,year=false,textwidth=1cm]{42}{\textcolor{blue}{SSAH \cite{jin2019ssah}}}

\chronoevent[markdepth=-15pt, year=false]{44}{\textbf{2020}}
\chronoevent[markdepth=-25pt,year=false,textwidth=1.2cm]{45}{\textcolor{cyan}{Bottleneck Autoencoder \cite{shen2020auto}}}
\chronoevent[markdepth=-125pt,year=false,textwidth=1cm]{48}{\textcolor{magenta}{SBS-CNN \cite{liu2020similarity}}}
\chronoevent[markdepth=-95pt,year=false,textwidth=1cm]{48}{\textcolor{magenta}{Pairwise \cite{wang2020deep1}}}
\chronoevent[markdepth=-63pt,year=false,textwidth=1.5cm]{48}{\textcolor{magenta}{Adversarial Siamese \cite{pandey2020stacked}}}
\chronoevent[markdepth=-25pt,year=false,textwidth=1cm]{50}{\textcolor{green}{Second-Order Similarity \cite{ng2020solar}}}

\chronoevent[markdepth=5pt,year=false,textwidth=1cm]{45}{\textcolor{red}{CNN Features \cite{gao2020multiple}}}
\chronoevent[markdepth=75pt,year=false,textwidth=1.5cm]{47}{\textcolor{blue}{BGAN+ \cite{song2020unified}}}
\chronoevent[markdepth=35pt,year=false,textwidth=1.5cm]{47}{\textcolor{blue}{Stack GAN \cite{pandey2020stacked}}}
\chronoevent[markdepth=5pt,year=false,textwidth=1.5cm]{50}{\textcolor{yellow}{Reinforcement Learning Based \cite{yang2020deep}}}

\stopchronology
\vspace{-2.2cm}
\caption{A chronological view of deep learning based image retrieval methods depicting the different type of neural networks used from 2011 to 2020. The convolutional neural network, autoencoder network, siamese \& triplet network, recurrent neural network, generative adversarial network, attention network and reinforcement learning network based deep learning approches for image retrieval are depicted in Red, Cyan, Magenta, Black, Blue, Green, and Yellow colors, respectively.}
\label{fig:network}
\end{figure*}

\subsection{Semi, Weakly, Pseudo and Self -supervised Approaches}
The semi-supervised approaches generally use a combination of labelled and unlabelled data for feature learning \cite{pan2015semi}, \cite{yan2017semi}.
Semi-supervised deep hashing (SSDH) \cite{zhang2017ssdh} uses labelled data for the empirical error minimization and both labelled and unlabelled data for embedding error minimization. The generative adversarial learning has been also utilized extensively in semi-supervised image retrieval \cite{wang2018semi}, \cite{jin2019ssah}, \cite{qiu2017deep}. A teacher-student based semi-supervised image retrieval \cite{zhang2019pairwise} uses the pairwise information learnt by the teacher network as the guidance to train the student network.

Weakly-supervised approaches have been also explored for the image retrieval. Tang et al. (2017) have put forward a weakly-supervised multimodal hashing (WMH) by utilizing the local discriminative and  geometric structures in the visual space \cite{tang2017weakly}.
Guan et al. (2018) \cite{guan2018tag} have performed the pre-training in weakly-supervised mode and fine-tuning in supervised mode. A weakly supervised deep hashing using tag embeddings (WDHT) \cite{gattupalli2019weakly} utilizes the word2vec semantic embeddings. A semantic guided hashing (SGH) \cite{li2020weakly} is used for image retrieval by simultaneously employing the weakly-supervised tag information and the inherent data relations.

The pseudo suervised networks have been also developed for image retrieval.
The pseudo triplets are utilized in \cite{gu2019unsupervised} for unsupervised image retrieval. K-means clustering based pseudo labels are generated and used for the training of a deep hashing network \cite{hu2017pseudo}, \cite{dong2020unsupervised}. An appealing performance has been observed using pseudo labels over CIFAR-10 and Flickr datasets for image retrieval. 

The self-supervision is another way of supervision used in some research works for image retrieval. For example, Li et al. (2018) \cite{li2018self} have used the adversarial networks in self-supervision mode 
by utilizing the multi-label annotations. Zhang et al. (2016) \cite{zhang2016play} have introduced a self-supervised temporal hashing (SSTH) for video retrieval. 

\subsection{Summary}
Following are the take aways from the above discussion on deep learning based models from the supervision perspective:
\begin{itemize}
    \item The supervised approaches utilize the class-specific semantic information through the classification error apart from the other objectives related to the hash code generation. Generally, the performance of supervised models is better than other models due to learning of the fine-grained and class specific information. 
    \item The unsupervised models make use of the unsupervised constraints on hash code (i.e., quantization loss, independent bits, etc.) and/or data reconstruction (i.e., using an autoencoder type of networks) to learn the features. 
    \item The semi-supervised approaches exploit the labelled and un-labelled data for the feature learning using deep networks. These approaches generally utilize the information from different modalities using different networks.
    \item The pseudo-supervised approaches generate the pseudo labels using some other methods to facilitate the training using generated labels. The self-supervised methods generate the temporal or generative information to learn the models over the training epochs.
    \item The minimal quantization error, independent bits, low dimensional feature, and discriminative code are the common objectives for most of the retrieval methods.
\end{itemize}

\section{Network Types For Image Retrieval} \label{network}
In this section, deep learning based image retrieval approaches are presented in terms of the different architectures.
A chronological overview from 2011 to 2020 is illustrated in Fig. \ref{fig:network} for different type of networks for image retrieval.

\subsection{Convolutional Neural Networks for Image Retrieval}
Convolutional neural networks (CNN) based feature learning has been utilized extensively for image retrieval as shown in Fig. \ref{fig:neural_code}. 
In 2014, CNN features off-the-shelf have shown a tremendous performance gain for image recognition and retrieval as compared to the hand-crafted features \cite{sharif2014cnn}. At the same time the activations of trained CNN has been also explored as the neural code for retrieval \cite{babenko2014neural}. An image representation learning has been also performed using the CNN model to generate the descriptor for image retrieval \cite{xia2014supervised}.
In 2016, pairwise labels are exploited to learn the CNN feature for image retrieval \cite{liu2016deep}, \cite{li2016feature}.
The CNN activations are heavily used to generate the hash codes for efficient image retrieval by employing the different losses \cite{gordo2016deep}, \cite{cao2016deep}, \cite{cao2017hashnet}.
The abstract features of CNN are learnt for the image retrieval in different modes, such as unsupervised image retrieval \cite{lin2016learning}, \cite{duan2017learning}, \cite{shen2018unsupervised}, \cite{lin2018unsupervised}, supervised image retrieval \cite{xia2014supervised}, \cite{liu2016deep}, \cite{lu2017deep}, \cite{kan2019supervised}, semi-supervised image retrieval \cite{zhang2017ssdh}, cross-modal retrieval \cite{wei2016cross}, \cite{yang2017pairwise}, sketch based image retrieval \cite{yu2018multi}, \cite{wang2019enhancing}, and object retrieval \cite{sun2014search}, \cite{gao2020multiple}.

\begin{figure}[!t]
    \centering
    \includegraphics[trim=10 330 10 10, clip, width=0.8\columnwidth]{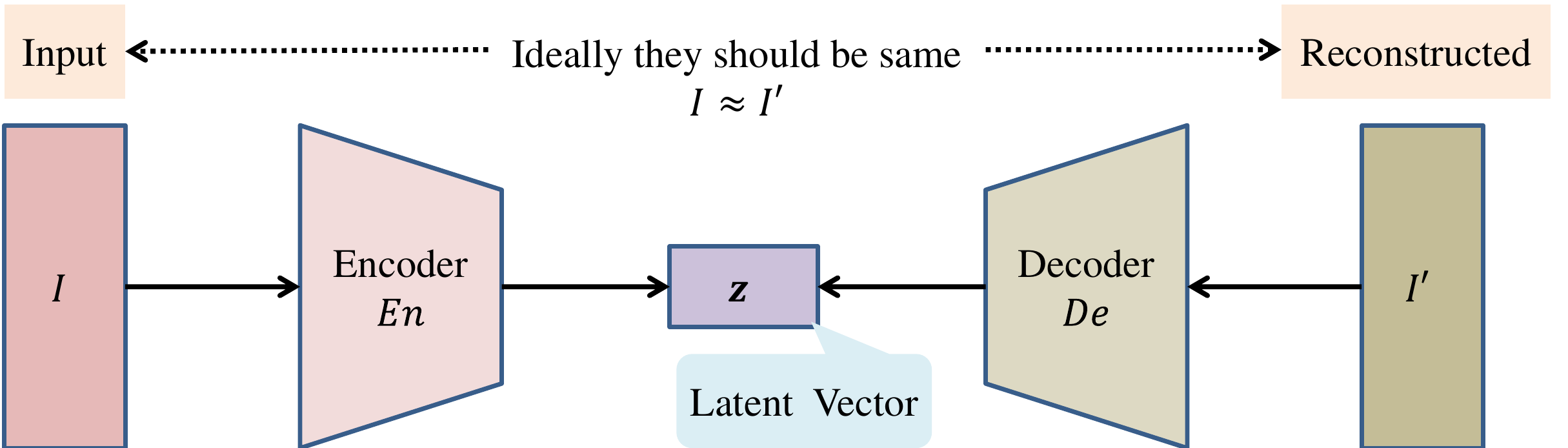}
    \caption{A typical Autoencoder network consisting of an Encoder and a Decoder network. Generally, the encoder is a CNN and the decoder is an up-CNN. The output of the encoder is a latent space which is used to generate the hash codes.}
    \label{fig:autoencoder}
\end{figure}

\subsection{Autoencoder Networks based Image Retrieval}
Autoencoder ($AE$) is a type of unsupervised neural network
which can be used to reconstruct the input image from the latent space as portrayed in Fig. \ref{fig:autoencoder}. Basically, it consists of two networks, namely encoder ($En$) and decoder ($De$). The encoder network transforms the input ($I$) into latent feature space ($z$) as $En: I \rightarrow z$. Whereas, the decoder network tries to reconstruct the original image ($I'$) from latent feature space as $De: z \rightarrow I'$. The model is trained by minimizing the reconstruction error between original image ($I$) and reconstructed image ($I'$) using $L_1$ or $L_2$ loss function.

The autoencoders have been very intensively used to learn the features as the latent space for image retrieval.
In the initial attempts, the deep autoencoder was used for image retrieval in 2011 \cite{krizhevsky2011using}.
A stacked denoising autoencoder is used to train the multiple deep neural networks for retrieval task \cite{wu2013online}.
Feng et al. (2014) have utilized the correspondence autoencoder (Corr-AE) for cross-modal retrieval \cite{feng2014cross}. 
A binary autoencoder is used to learn the binary code for fast image retrieval by reconstructing the image from that binary code function \cite{carreira2015hashing}. The use of autoencoder in image retrieval has witnessed a huge progressed in recent years, such as
Deep variational binaries (DVB) using the variational Bayesian networks \cite{shen2019unsupervised}; 
Autoencoder over the triplet  \cite{gu2019unsupervised}; and Relaxed binary autoencoder (RBA) \cite{do2019simultaneous} are investigated in 2019.
In a recent work, double latent bottlenecks is used in autoencoder \cite{shen2020auto}. It includes binary latent variable and continuous latent variable. The latent variable bottleneck exchanges crucial information collaboratively and the binary codes bottleneck uses a code-driven graph to capture the intrinsic data structure.

\begin{figure}[!t]
    \centering
    \includegraphics[trim=10 215 90 10, clip, width=0.8\columnwidth]{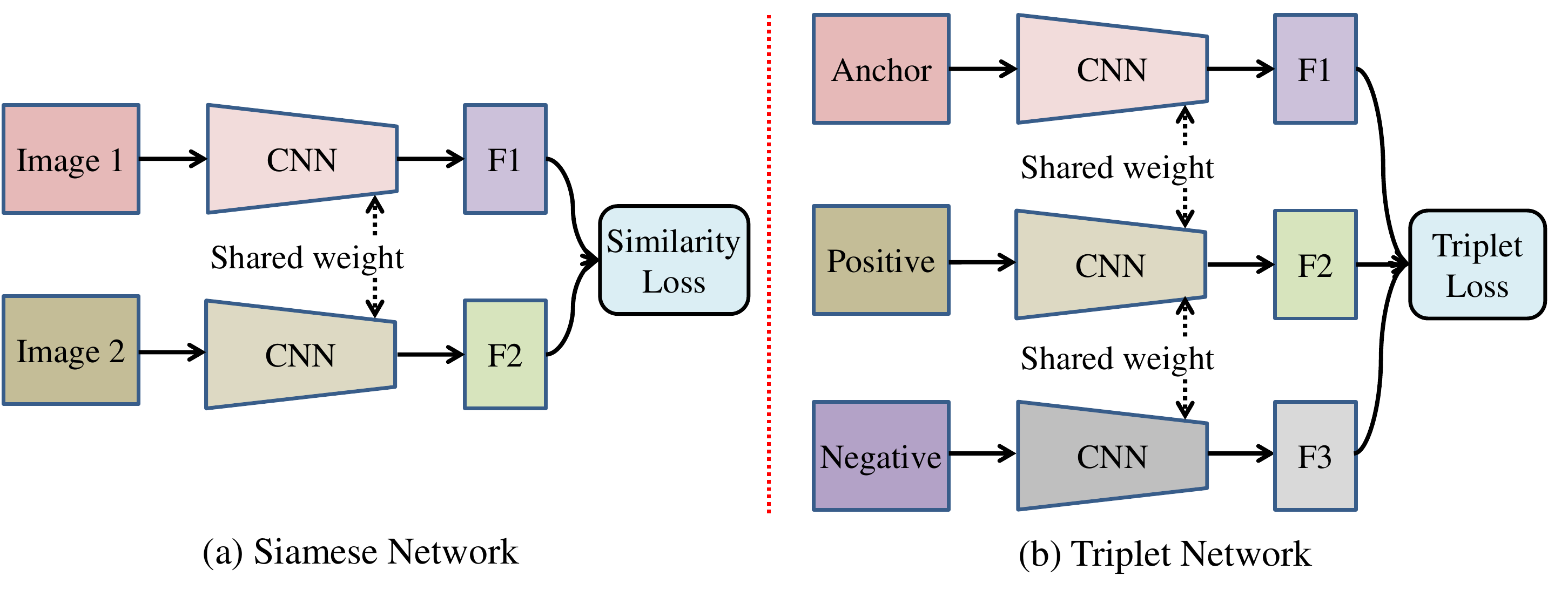}
    \caption{(a) Siamese network computes the similarity between image pairs. (b) Triplet network minimizes the distance between the anchor and positive and maximizes the distance between the anchor and negative in feature space.}
    \label{fig:siamese}
\end{figure}

\subsection{Siamese and Triplet Networks for Image Retrieval}
\subsubsection{Siamese Network}
The siamese is type of neural network
that exploits the distance between features of image pairs as depicted in Fig. \ref{fig:siamese}(a). 
The siamese network based learnt features have shown very promising performance for fine-grained image retrieval \cite{wang2014learning}. A pair of similar or dissimilar images is jointly processed by Liu et al. \cite{liu2016deep} to produce 1 or -1 output by CNN to learn the feature for image retrieval. 
Ong et al. (2017) have used the fisher vector computed on top of the CNN feature in autoencoder network to generate the discriminating feature descriptor for image retrieval \cite{ong2017siamese}.
The siamese network is also used to develop the light weight models for efficient image retrieval \cite{jose2017binary}, \cite{liu2020similarity}.
A pairwise similarity-preserving quantization loss is employed in \cite{wang2020deep1}.
The siamese network is used with the stacked adversarial network in \cite{pandey2020stacked}.
The siamese network is also used for patch based image matching \cite{zagoruyko2015learning}, \cite{han2015matchnet}.

\subsubsection{Triplet Network}
A triplet network is a variation of siamese network 
which utilizes a triplet of images, including an anchor, a positive and a negative image as shown in Fig. \ref{fig:siamese}(b). The triplet network minimizes the distance between the features of anchor and positive image and maximizes the distance between the features of anchor and negative image, simultaneously. 
In 2015, a triplet ranking loss is utilized on top of the shared CNN features to learn the network for computation of binary descriptors for image retrieval \cite{lai2015simultaneous}.
An adjacency consistency based regularization term is introduced in the triplet network to enforce the discriminative ability of the CNN feature description \cite{zhang2015bit}.
Zhuang et al. (2016) \cite{zhuang2016fast} have used triplet to learn the hash code by employing the relation weights matrix and graph cuts optimization. 
The triplet ranking loss, orthogonality constraint and softmax loss are minimized jointly in \cite{yao2016deep}.
Triplet based siamese networks are als used for image rerieval \cite{kumar2016learning}, \cite{gordo2017end}. Triplet quantization based objective function minimizes the information loss  \cite{zhou2017deep}, \cite{liu2018deep}.
The triplet based feature learning has been also exploited for sketch based image retrieval \cite{dey2019doodle}. Triplets are also exploited for supervised hashing \cite{wang2016deep} and unsupervised hashing \cite{huang2017unsupervised}, \cite{gu2019unsupervised} for image retrieval.

\subsection{Generative Adversarial Networks based Retrieval}
The generative adversarial network (GAN)
uses two networks, i.e., generator and discriminator. The generator network generates the new samples in the training set from the random vector. Whereas, the discriminator network distinguishes between generated image and original image. 
In 2018, Song et al. have introduced a binary generative adversarial network (BGAN) 
for generating the representational binary codes for image retrieval \cite{song2018binary}. 
At the same time, a regularized GAN is used to introduce the BinGAN model \cite{zieba2018bingan} to learn the compact binary patterns. The BinGAN uses two regularizers, including a distance matching regularizer and a binarization representation entropy (BRE) regularizer. 
In 2018, the generative networks are also utilized in \cite{ghasedi2018unsupervised} to develop HashGAN in an unsupervised manner to generate the hash code for image retrieval. At the same time another HashGAN is developed by employing the paired conditional Wasserstein GAN for image retrieval \cite{cao2018hashgan}. 
GAN has been also used for cross-modal retrieval \cite{wang2017adversarial}, \cite{zhang2018attention}, \cite{li2018self}, \cite{gu2019adversary}, semi-supervised hashing \cite{wang2018semi}, \cite{jin2019ssah}, sketch based image retrieval \cite{zhang2018generative}, \cite{kumar2019generative}, \cite{pandey2020stacked} and unsupervised adversarial hashing \cite{deng2019unsupervised}. 
In 2020, binary generative adversarial networks based unified BGAN+ framework \cite{song2020unified} is developed for image retrieval.

\subsection{Attention Networks for Image Retrieval}
The attention has been observed as a very effective way of modelling the saliency information into the feature space to avoid the effect of background.
In 2017, Noh et al. have used the attention-based keypoints to select the important deep local features \cite{noh2017large}. 
Yang et al. (2017) have introduced a two-stream attentive CNNs by fusing a Main and an Auxiliary CNN (MAC) for image retrieval \cite{yang2017two}. The main CNN focuses over the discriminative visual features for semantic information, whereas the auxiliary CNN focuses over the part of features for attentive information.
Similarily, two sub-networks are employed in \cite{ge2019deep} for spatial attention and global features, respectively. 
Recently, Ng et al. \cite{ng2020solar} have computed the second-order similarity (SOS) loss over the attention based selected regions of the input image for image retrieval. 
The attention based models are developed for cross-modal retrieval \cite{zhang2018attention} and fine-grained sketch-based image retrieval \cite{song2017deep}. The gradient attention network based deep hashing \cite{huang2019accelerate} enforces the CNN binary features of a pair to minimize the distances between them, irrespective of their signs or directions. In order to localize the important image region for the feature description, an attentional heterogeneous bilinear network is employed in \cite{su2020look} for fashion image retrieval.

\begin{figure*}[!t]
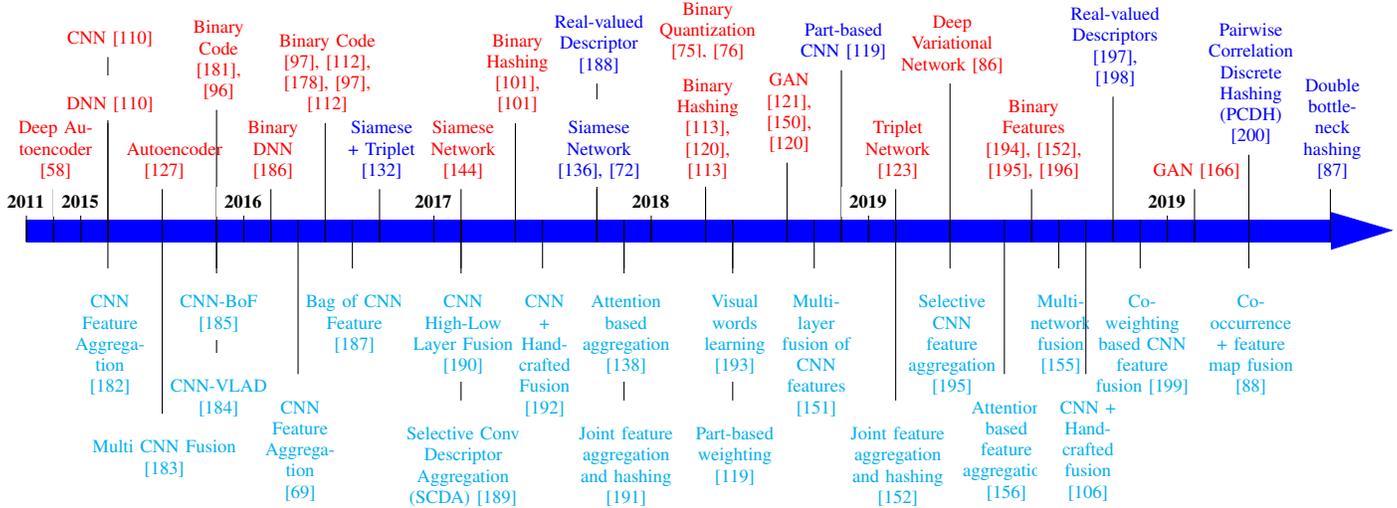

\centering
\vspace{-4.0cm}
\startchronology[startyear=1,stopyear=49, startdate=false, color=blue!100, stopdate=false, arrow=true, arrowwidth=0.8cm, arrowheight=0.5cm]
\setupchronoevent{textstyle=\scriptsize,datestyle=\scriptsize}
\chronoevent[markdepth=-15pt, year=false]{1}{\textbf{2011}}
\chronoevent[markdepth=-25pt,year=false,textwidth=1cm]{2}{\textcolor{red}{Deep Autoencoder \cite{krizhevsky2011using}}}
\chronoevent[markdepth=-15pt, year=false]{3}{\textbf{2015}}
\chronoevent[markdepth=-75pt,year=false,textwidth=1.5cm]{4}{\textcolor{red}{CNN \cite{erin2015deep}}}
\chronoevent[markdepth=-50pt,year=false,textwidth=1.5cm]{4}{\textcolor{red}{DNN \cite{erin2015deep}}}
\chronoevent[markdepth=-25pt,year=false,textwidth=1cm]{6}{\textcolor{red}{Autoencoder \cite{carreira2015hashing}}}
\chronoevent[markdepth=-55pt,year=false,textwidth=1cm]{8}{\textcolor{red}{Binary Code \cite{lin2015rapid}, \cite{shen2015supervised}}}
\chronoevent[markdepth=10pt,year=false,textwidth=1cm]{4}{\textcolor{cyan}{CNN Feature Aggregation \cite{babenko2015aggregating}}}
\chronoevent[markdepth=65pt,year=false,textwidth=2cm]{6}{\textcolor{cyan}{Multi CNN Fusion \cite{liu2015deepindex}}}
\chronoevent[markdepth=42pt,year=false,textwidth=1.3cm]{8}{\textcolor{cyan}{CNN-VLAD \cite{yue2015exploiting}}}
\chronoevent[markdepth=10pt,year=false,textwidth=1.2cm]{8}{\textcolor{cyan}{CNN-BoF \cite{uricchio2015fisher}}}
\chronoevent[markdepth=-15pt, year=false]{9}{\textbf{2016}}
\chronoevent[markdepth=-25pt,year=false,textwidth=1cm]{10}{\textcolor{red}{Binary DNN \cite{do2016learning}}}
\chronoevent[markdepth=-50pt,year=false,textwidth=1.5cm]{12}{\textcolor{red}{Binary Code \cite{liu2016deep}, \cite{lin2016learning}, \cite{zhang2016play}, \cite{liu2016deep}, \cite{lin2016learning}}}
\chronoevent[markdepth=-25pt,year=false,textwidth=1cm]{14}{\textcolor{blue}{Siamese + Triplet \cite{kumar2016learning}}}
\chronoevent[markdepth=50pt,year=false,textwidth=1cm]{11}{\textcolor{cyan}{CNN Feature Aggregation \cite{husain2016improving}}}
\chronoevent[markdepth=10pt,year=false,textwidth=1.3cm]{13}{\textcolor{cyan}{Bag of CNN Feature \cite{mohedano2016bags}}}
\chronoevent[markdepth=-15pt, year=false]{16}{\textbf{2017}}
\chronoevent[markdepth=-25pt,year=false,textwidth=1cm]{17}{\textcolor{red}{Siamese Network \cite{jose2017binary}}}
\chronoevent[markdepth=-50pt,year=false,textwidth=1cm]{19}{\textcolor{red}{Binary Hashing \cite{li2017deep}, \cite{li2017deep}}}
\chronoevent[markdepth=-65pt,year=false,textwidth=1.2cm]{22}{\textcolor{blue}{Real-valued Descriptor \cite{qayyum2017medical}}}
\chronoevent[markdepth=-25pt,year=false,textwidth=1.2cm]{22}{\textcolor{blue}{Siamese Network \cite{ong2017siamese}, \cite{gordo2017end}}}
\chronoevent[markdepth=60pt,year=false,textwidth=1.5cm]{17}{\textcolor{cyan}{Selective Conv Descriptor Aggregation (SCDA) \cite{wei2017selective}}}
\chronoevent[markdepth=10pt,year=false,textwidth=1.5cm]{17}{\textcolor{cyan}{CNN High-Low Layer Fusion \cite{yu2017exploiting}}}
\chronoevent[markdepth=60pt,year=false,textwidth=1.5cm]{23}{\textcolor{cyan}{Joint feature aggregation and hashing \cite{do2017simultaneous}}}
\chronoevent[markdepth=10pt,year=false,textwidth=1.5cm]{23}{\textcolor{cyan}{Attention based aggregation \cite{yang2017two}}}
\chronoevent[markdepth=10pt,year=false,textwidth=0.7cm]{20}{\textcolor{cyan}{CNN + Hand-crafted Fusion \cite{zhou2017collaborative}}}
\chronoevent[markdepth=-15pt, year=false]{24}{\textbf{2018}}
\chronoevent[markdepth=-70pt,year=false,textwidth=1.5cm]{26}{\textcolor{red}{Binary Quantization \cite{cao2018deep}, \cite{su2018greedy}}}
\chronoevent[markdepth=-25pt,year=false,textwidth=1.2cm]{26}{\textcolor{red}{Binary Hashing \cite{lin2018unsupervised}, \cite{ghasedi2018unsupervised}, \cite{lin2018unsupervised}}}
\chronoevent[markdepth=-35pt,year=false,textwidth=1cm]{29}{\textcolor{red}{GAN \cite{song2018binary}, \cite{zieba2018bingan}, \cite{ghasedi2018unsupervised}}}
\chronoevent[markdepth=-70pt,year=false,textwidth=1.5cm]{31}{\textcolor{blue}{Part-based CNN \cite{xu2018unsupervised}}}
\chronoevent[markdepth=60pt,year=false,textwidth=1.2cm]{27}{\textcolor{cyan}{Part-based weighting \cite{xu2018unsupervised}}}
\chronoevent[markdepth=10pt,year=false,textwidth=1.2cm]{27}{\textcolor{cyan}{Visual words learning \cite{liu2019e2bows}}}
\chronoevent[markdepth=10pt,year=false,textwidth=1cm]{30}{\textcolor{cyan}{Multi-layer fusion of CNN features \cite{yu2018multi}}}
\chronoevent[markdepth=-15pt, year=false]{32}{\textbf{2019}}
\chronoevent[markdepth=-25pt,year=false,textwidth=1cm]{33}{\textcolor{red}{Triplet Network \cite{gu2019unsupervised}}}
\chronoevent[markdepth=-65pt,year=false,textwidth=1.5cm]{35}{\textcolor{red}{Deep Variational Network \cite{shen2019unsupervised}}}
\chronoevent[markdepth=-25pt,year=false,textwidth=1.3cm]{38}{\textcolor{red}{Binary Features \cite{do2019compact}, \cite{do2019simultaneous}, \cite{do2019selective}, \cite{lu2019discrete}}}
\chronoevent[markdepth=-60pt,year=false,textwidth=1.2cm]{41}{\textcolor{blue}{Real-valued Descriptors \cite{dubey2019local}, \cite{ji2019deep}}}
\chronoevent[markdepth=60pt,year=false,textwidth=1.5cm]{33}{\textcolor{cyan}{Joint feature aggregation and hashing \cite{do2019simultaneous}}}
\chronoevent[markdepth=10pt,year=false,textwidth=1.2cm]{35}{\textcolor{cyan}{Selective CNN feature aggregation \cite{do2019selective}}}
\chronoevent[markdepth=50pt,year=false,textwidth=1.2cm]{37}{\textcolor{cyan}{Attention based feature aggregation \cite{wei2019saliency}}}
\chronoevent[markdepth=10pt,year=false,textwidth=1.2cm]{39}{\textcolor{cyan}{Multi-network fusion \cite{ge2019deep}}}
\chronoevent[markdepth=10pt,year=false,textwidth=1.25cm]{42}{\textcolor{cyan}{Co-weighting based CNN feature fusion \cite{zhu2019co}}}
\chronoevent[markdepth=50pt,year=false,textwidth=1.2cm]{40}{\textcolor{cyan}{CNN + Hand-crafted fusion \cite{kan2019supervised}}}
\chronoevent[markdepth=-15pt, year=false]{43}{\textbf{2019}}
\chronoevent[markdepth=-25pt,year=false,textwidth=1.2cm]{44}{\textcolor{red}{GAN \cite{song2020unified}}}
\chronoevent[markdepth=-38pt,year=false,textwidth=1.3cm]{46}{\textcolor{blue}{Pairwise Correlation Discrete Hashing (PCDH) \cite{chen2020deep}}}
\chronoevent[markdepth=-25pt,year=false,textwidth=1cm]{49}{\textcolor{blue}{Double bottleneck hashing \cite{shen2020auto}}}
\chronoevent[markdepth=10pt,year=false,textwidth=1.2cm]{46}{\textcolor{cyan}{Co-occurrence + feature map fusion \cite{forcen2020co}}}
\stopchronology
\vspace{-2.7cm}
\caption{A chronological view of deep learning based image retrieval methods depicting the different type of descriptors. The binary and real-valued feature vector based models are presented in Red and Blue colors, respectively. The feature aggregation based models are presented in Cyan color.}
\label{fig:descriptor}
\end{figure*}

\subsection{Recurrent Neural Networks for Image Retrieval}
In 2018, Lu et al. have utilized the recurrent neural network (RNN) concept to perform a hierarchical recurrent neural hashing (HRNH) to produce the effective hash codes for image retrieval \cite{lu2017hierarchical}.
In 2017, Shen et al. have used the region-based convolutional networks with long short-term memory (LSTM) modules for textual-visual cross retrieval \cite{shen2017deep}. 
Bai et al. (2019) have also employed the LSTM based recurrent deep network in the triplet hashing framework to naturally inherit the useful information for image retrieval \cite{bai2019deep}.

\subsection{Reinforcement Learning Networks based Retrieval}
In 2018, Yuan et al. have exploited the reinforcement learning for image retrieval \cite{yuan2018relaxation}. They have used a relaxation free method through policy gradient to generate the hash codes for image retrieval. The similarity preservation via the generated binary codes is used as the reward function. 
In 2020, Yang et al. \cite{yang2020deep} have utilized the deep reinforcement learning to perform the de-redundancy in hash bits to get rid of redundant and/or harmful bits, which reduces the ambiguity in the similarity computation for image retrieval.

\subsection{Summary}
The summary of the different network driven deep learning based image retrieval approaches is as follows:
\begin{itemize}
    \item The convolutional neural network features are exploited for the hash code and descriptor learning by employing the various constraints like classification error, quantization error, independent bits, etc.
    \item In order to make the features more representative of the image, the autoencoder networks are used which enforces the learning based on the reconstruction loss.
    \item The discriminative power of descriptive hash code is enhanced by exploiting the siamese and triplet networks. Different constraints are used on the hash code to make it discriminative and compact.
    \item The generative adversarial network based approaches have been highly utilized to improve the discriminative ability and robustness of the learnt features by encoder network guided through the discriminator network.
    \item The automatic important feature selection is performed using attention module to control the redundancy in the feature space. The recurrent neural network and reinforcement learning network have been also shown very effective for the image retrieval.
\end{itemize}

\section{Type of Descriptors for Image Retrieval} \label{descriptor}
This section covers the binary hash codes for efficient image retrieval, real-valued descriptors and feature aggregation for discriminative image retrieval as depicted in Fig. \ref{fig:descriptor}. 

\subsection{Binary Descriptors}
Different types of networks are used to learn the binary description such as deep neural networks \cite{erin2015deep}, convolutional neural networks \cite{lin2015deep}, autoencoder networks \cite{carreira2015hashing}, siamese networks \cite{jose2017binary}, triplet networks \cite{gu2019unsupervised}, generative adversarial networks, \cite{song2020unified}, and variational networks \cite{shen2019unsupervised}.
In 2015, Liong et al. \cite{erin2015deep} have introduced a supervised deep hashing (SDH). The SDH method uses the quantization loss, balanced bits and independent bits constraints. Binary hash code is also learnt through a latent layer in a supervised manner in \cite{lin2015deep}.
A binary autoencoder  \cite{krizhevsky2011using},  \cite{carreira2015hashing} and a siamese network \cite{jose2017binary} are used to learn the binary features for efficient image retrieval.
A binary deep neural network (BDNN) is proposed by converting a hidden layer output to binary code \cite{do2016learning}, \cite{do2019compact}. 
The binary code is jointly learnt with feature aggregation in \cite{do2019simultaneous}. 
A masking technique over the convolutional features is used to generate the binary description for image retrieval \cite{do2019selective}.
A ranking optimization discrete hashing (RODH) approach is used in \cite{lu2019discrete} by generating the discrete hash codes (+1 or -1) by employing the ranking information. 
A cauchy quantization loss is used in \cite{cao2018deep} to improve the discriminative power of binary descriptors. An iterative quantization approach is used to convert the features into binary codes to avoid the quantization loss \cite{su2018greedy}. 
Binary hash code is also used for clothing image retrieval \cite{lin2015rapid}. 
The binary description is learnt through the supervised \cite{shen2015supervised}, \cite{liu2016deep}, \cite{li2017deep}, unsupervised \cite{lin2016learning}, \cite{lin2018unsupervised}, \cite{ghasedi2018unsupervised} and self-supervised \cite{zhang2016play} deep learning techniques. 
Among the generative approaches, a binary generative adversarial network (BGAN) is used to learn the binary code \cite{song2018binary}. At the same time a regularized GAN is used by maximizing the entropy of binarized layer for image retrieval \cite{zieba2018bingan}. The GAN is trained in unsupervised mode \cite{ghasedi2018unsupervised} to learn the binary codes for image retrieval. In 2020, the binary GAN \cite{song2020unified} is used for image retrieval and compression jointly.

\subsection{Real-Valued Descriptors}
The binary hashing approaches have the obvious shortcomings. First, it is difficult to represent the fine-grained similarity using binary code. Second, the generation of similar binary codes is common even for different images. Thus, researchers have also used the real-valued features to represent the images for the retrieval.
The siamese networks have been extensively used to learn the real-valued feature descriptor for image retrieval \cite{kumar2016learning}, \cite{ong2017siamese}, \cite{gordo2017end}. 
In 2018, part-based CNN features are utilized to extract a non-binary hash code \cite{xu2018unsupervised}.
The real-valued descriptors generated using CNNs are used for medical image retrieval \cite{qayyum2017medical}, \cite{dubey2019local} and cross-modal retrieval \cite{ji2019deep}.
Chen et al. (2020) \cite{chen2020deep} have developed a pairwise correlation discrete hashing (PCDH) by exploiting the pairwise correlation of deep features for image retrieval. Shen et al. (2020) \cite{shen2020auto} have also used the real-valued descriptors with the help of double bottleneck hashing approach for image retrieval.

\subsection{Aggregation of Descriptors}
Several researchers have also tried to combine/fuse the feature at different stages of the network or multiple networks to generate the aggregation of descriptors for image retrieval \cite{husain2016improving}, \cite{babenko2015aggregating}, \cite{uricchio2015fisher}. 
Different strategies have been excercised for aggregation of features, such as vector locally aggregated descriptors (VLAD) \cite{yue2015exploiting} on the features extracted from different layers; 
bag of local convolutional features \cite{mohedano2016bags}; 
selective convolutional descriptor aggregation \cite{wei2017selective}, \cite{do2019selective}; fusion of multi-layer features \cite{yu2017exploiting}, \cite{yu2018multi}; part-based weighting aggregation \cite{xu2018unsupervised}; joint training of feature aggregation and hashing \cite{do2017simultaneous}; learning of feature aggregation and hash function in a joint manner \cite{do2019simultaneous}; and co-weighting based CNN feature fusion \cite{zhu2019co}. 
The features from different CNNs are also integrated for image retrieval \cite{liu2015deepindex}, \cite{yang2017two}, \cite{wei2019saliency}. 
One main sub-network and other attention-based sub-network are also fused at the last fully connected layer in \cite{ge2019deep}. 
The hand-designed features are fused with CNNs \cite{kan2019supervised}, \cite{zhou2017collaborative}. 
Recently, Forcen et al. (2020) \cite{forcen2020co} have generated the image representation by combining a co-occurrence map with the feature map for image retrieval.

\subsection{Summary}
The followings are the summary of deep learning based approaches from the perspective of the type of feature descriptor:
\begin{itemize}
    \item In order to facilitate the large-scale image retrieval, the compact and binary hash codes are generated using different networks. Different methods try to improve the discriminative ability, lower redundancy among bits, generalization of the binary hash code, etc. in different supervision modes.
    \item The real-valued descriptors concentrate over the discriminative ability of the learnt features for image retrieval at the cost of increased computational complexity for feature matching. Such methods try to increase the robustness and reduce the dimensionality of the descriptors.
    \item Feature aggregation approaches try to utilize the complementary information between the features of different networks, the features of different sub-network, and the features of different layers of same network to improve the image retrieval performance.
\end{itemize}

\section{Retrieval Type} \label{retrieval}
Various retrieval types have been explored using deep learning approaches based on the nature of the problem and data as discussed in this section. 

\subsection{Cross-modal Retrieval}
The cross-modal retrieval refers to the image retrieval involving more than one modality by measuring the similarity between heterogeneous data objects. 
Feng et al. (2014) have introduced a correspondence autoencoder (Corr-AE) network for cross-modal retrieval \cite{feng2014cross}.
In 2016 \cite{cao2016deepcross}, a deep visual-semantic hashing (DVSH) network is developed for sentence and image based cross-modal retrieval by jointly learning the embeddings for images and sentences. Textual-visual deep binaries (TVDB) model represents the long descriptive sentences along with its corresponding informative images \cite{shen2017deep}.
The CNN visual features have been also exploited for cross-modal retrieval, such as CNN off-the-shelf features for labelled annotation \cite{wei2016cross}, CNN features with bi-directional hinge loss \cite{wu2017online}, and pairwise constraints based deep hashing network \cite{yang2017pairwise}.
The adversarial neural network is also employed for cross-modal retrieval, such as adversarial cross-modal retrieval (ACMR) \cite{wang2017adversarial}, self-supervised adversarial hashing (SSAH) \cite{li2018self}, attention-aware deep adversarial hashing (ADAH) \cite{zhang2018attention}, adversary guided asymmetric hashing (AGAH) \cite{gu2019adversary}, deep multi-level semantic hashing (DMSH) \cite{ji2019deep}, and teacher-student learning \cite{liu2020improving}.

\subsection{Sketch Based Image Retrieval}
Sketch based image retrieval (SBIR) is a special case of cross-modal retrieval where the query image is in the sketch domain the retrieval has to be performed in the image domain \cite{lei2019semi}. 
In 2017, a fine-grained SBIR (FG-SBIR) \cite{song2017deep} is explored with the help of attention module and higher-order learnable energy function loss. 
Liu et al. (2017) \cite{liu2017deep} have introduced a semi-heterogeneous deep sketch hashing (DSH) model for SBIR by utilizing the representation of free-hand sketches.
The sketches and natural photos are mapped in multiple layers in a deep CNN framework in \cite{yu2018multi} for SBIR. A zero-shot SBIR (ZS-SBIR) is proposed for retrieval of photos from unseen categories \cite{dey2019doodle}.
Wang et al. (2019) \cite{wang2019enhancing} have proposed a CNN based SBIR re-ranking approach to refine the retrieval results. 
The generative adversarial networks have been also exploited extensively for SBIR, such as generative domain-migration hashing (GDH) using cycle consistency loss \cite{zhang2018generative}, class sketch conditioned generative model \cite{kumar2019generative}, semantically aligned paired cycle-consistent generative model \cite{dutta2019semantically}, and stacked adversarial network \cite{pandey2020stacked}.

\subsection{Multi-label Image Retrieval}
Multi-label retrieval involves multiple categorical labels while generating the image representations for image retrieval. Several deep learning approaches have been investigated for multi-label image retrieval using different strategies, such as multilevel similarity information \cite{zhao2015deep}, multilevel semantic similarity preserving hashing \cite{wu2017deep}, multi-label annotations \cite{li2018self}, category-aware object based hashing \cite{lai2016instance}, \cite{chen2020multiple}, and fine-grained features for multilevel similarity hashing \cite{qin2020deep}. 
Readers may refer to the survey of multi-label image retrieval \cite{rodrigues2020deep} published in 2020 for wider aspects and developments.

\subsection{Instance Retrieval}
In 2015, Razavian et al. have developed a baseline for deep CNN based visual instance retrieval \cite{sharif2015baseline}. 
An instance-aware image representations for multi-label image data by modeling the features of one category in a group is proposed in \cite{lai2016instance}. Other approaches for image instance retrieval includes bags of local convolutional features \cite{mohedano2016bags}, learning global representations \cite{gordo2016deep}, and group invariant deep representation \cite{morere2017group}. 
In 2020, Chen et al. have proposed a deep multiple-instance ranking based hashing (DMIRH) model for multi-label image retrieval by employing the category-aware bag of feature \cite{chen2020multiple}. 
More details about image instance retrieval can be found in the survey compiled in \cite{zheng2017sift}.

\subsection{Object Retrieval}
The object retrieval aims to perform the retrieval based on the features derived from the specific objects in the image.
In 2014, Sun et al. have extracted the CNN features from the region of interest detected through object detection technique for object based retrieval \cite{sun2014search}. Several deep learning models have been investigated for object retrieval, such as integral image driven max-pooling on CNN activations  \cite{tolias2015particular}, pooling of the relevant features based on the region proposal network \cite{gordo2016deep}, replicator equation based simultaneous selection and weighting of the primitive deep CNN features \cite{pang2018building}, co-weighting based aggregation of the semantic CNN features \cite{zhu2019co}, and consideration of spatial and channel contribution to improve the region detection \cite{shi2019exploring}. 
Gao et al. (2020) \cite{gao2020multiple} have performed the 3D object retrieval with the help of a multi-view discrimination and pairwise CNN (MDPCNN) network.

\subsection{Semantic Retrieval}
In 2016, Yao et al. \cite{yao2016deep} have introduced a deep semantic preserving and ranking-based hashing (DSRH) method by exploiting the hash and classification losses. Similar losses are also used in \cite{guo2016hash}. A deep visual-semantic quantization (DVSQ) \cite{cao2017deep} is used by  jointly learning the visual-semantic embeddings and quantizers.
An adaptive Gaussian filter based aggregation of CNN features is used in \cite{zhu2019co} to exploit the semantic information.
Semantic hashing has been also extensively performed for sketch based image retrieval \cite{song2017deep}, \cite{dey2019doodle}, \cite{dutta2019semantically}, \cite{wang2019enhancing}, cross-modal retrieval \cite{ji2019deep}, \cite{cao2016deepcross}, \cite{shen2017deep}. Other notable deep learning based works that model the semantic information include Multi-label retrieval \cite{zhao2015deep}, unsupervised image retrieval \cite{qin2020unsupervised}, supervised image retrieval \cite{yang2017supervised}, and semi-supervised image retrieval \cite{zhang2017ssdh}.
Semantic similarity in Hamming space based deep position-aware hashing (DPAH) \cite{wang2020deep} and semantic affinity deep semantic reconstruction hashing (DSRH) \cite{wang2020deep1} are the recent methods for semantic retrieval.

\subsection{Fine-Grained Image Retrieval}
In order to increase the discriminative ability of the deep learnt descriptors, many researchers have utilized the fine-grained constraints in deep networks. Different works have incorporated the fine-grained property using different approaches, such as  capturing the inter-class and intra-class image similarities using a siamese network \cite{wang2014learning}, attention modules based incorporation of the spatial-semantic information \cite{song2017deep}, using selective CNN features \cite{wei2017selective}, fine-grained ranking using the weighted Hamming distance \cite{zhang2018query}, 
using the multilevel semantic similarity between multi-label image pairs \cite{qin2020deep}, and using a piecewise cross entropy loss \cite{zeng2020fine}.

\subsection{Asymmetric Quantization based Retrieval}
In 2017, Wu et al. have performed the online asymmetric similarity learning to preserve the similarity between heterogeneous data \cite{wu2017online}.
An asymmetric deep supervised hashing (ADSH) is used by learning the deep hash function only for query images, while the hash codes for gallery images are directly learned \cite{jiang2018asymmetric}.
In 2019, Yang et al. have investigated an asymmetric deep semantic quantization (ADSQ) using three stream networks to model the heterogeneous data \cite{yang2019asymmetric}.
A similarity preserving deep asymmetric quantization (SPDAQ) is proposed by exploiting the image subset and the label information of all the database items \cite{chen2019similarity}.
An adversary guided asymmetric hashing (AGAH) is introduced in \cite{gu2019adversary} with the help of adversarial learning guided multi-label attention module for cross-modal image retrieval.

\subsection{Summary}
Based on the progress in image retrieval using deep learning methods for different retrieval types, following are the outlines drawn from this section:
\begin{itemize}
    \item The cross-modal retrieval approaches learn the joint features for multiple modality using different networks. The recent methods utilize of the adversarial network for cross-modal retrieval. The similar observation and trend has been also witnessed for sketch based image retrieval.
    \item The multi-label and instance retrieval approaches are generally useful where more than one type of visual scenarios is present in the image. The deep learning based approaches are able to handle such retrieval by facilitating the feature learning through different type of networks.
    \item The region proposal network based feature selection has been employed by the existing deep learning methods for the object retrieval. 
    \item The semantic information of the image has been used by different networks through abstract features to enhance the semantic image retrieval. The reconstruction based network is more suitable for semantic preserving hashing.
    \item Different feature selection and aggregation based networks have been utilized for fine-grained image retrieval. 
    \item The asymmetric hashing has also shown the suitability of deep learning models by processing the query and gallery images with different networks.
\end{itemize}

\section{Miscellaneous} \label{misc}
This section covers the deep learning models for retrieval in terms of the different losses, applications and other aspects.

\subsection{Progress in Retrieval Loss}
A siamese based loss function is used in \cite{kumar2016learning} by Kumar et al. (2016) for minimizing the global loss leading to discriminative feature learning.
Zhou et al. (2017) have used the triplet quantization loss for deep hashing, which is based on the similarity between the anchor-positive pairs and anchor-negative pairs \cite{zhou2017deep}.
A listwise loss has been employed by Revaud et al. in 2019 \cite{revaud2019learning} to directly optimize the global mean average precision in end-to-end deep learning. 
In 2020, a piecewise cross entropy loss function is used in \cite{zeng2020fine} for fine-grained image retrieval.
Several innovative losses have been used by the different feature learning approaches such as a lifted structured loss \cite{oh2016deep} and ranking loss \cite{wang2018semi}.

\begin{table}[!t]
    \centering
    \caption{Mean Average Precision (mAP) with 5000 retrieved images (mAP@5000) in \% for different deep learning based image retrieval approaches over NUS-WIDE, MS COCO and CIFAR-10 datasets. Note that $2^{nd}$ column list the reference from where the results of corresponding approach are considered. Followings are the used acronyms for different network types in the results: DNN - Deep Neural Network, CNN - Convolutional Neural Network, SN - Siamese Network, TN - Triplet Network, GAN - Generative Adversarial Network, DQN - Deep Q Network, PTN - Parametric Transformation Network, DVN - Deep Variational Networks, and AE - Autoencoder.}
    \begin{tabular}{|m{0.13\textwidth}|m{0.022\textwidth}|m{0.022\textwidth}|m{0.017\textwidth}|m{0.017\textwidth}|m{0.017\textwidth}|m{0.017\textwidth}|m{0.017\textwidth}|m{0.017\textwidth}|}
    \hline
     &  &  & \multicolumn{3}{c|}{NUS-WIDE} & \multicolumn{3}{c|}{MS COCO}\\
    \cline{4-9}
    Method Name & Net. Type & Result Source & 16 Bits & 32 Bits & 64 Bits & 16 Bits & 32 Bits & 64 Bits \\ 
    \hline
    CNNH'14 \cite{xia2014supervised} & CNN & \cite{cao2017hashnet} & 57.0 & 58.3 & 60.0 & 56.4 & 57.4 & 56.7 \\\hline
    SDH'15 \cite{shen2015supervised} & PTN & \cite{cao2017hashnet} & 47.6 & 55.5 & 58.1 & 55.5 & 56.4 & 58.0 \\\hline
    DNNH'15 \cite{lai2015simultaneous} & DNN & \cite{cao2017hashnet} & 59.8 & 61.6 & 63.9 & 59.3 & 60.3 & 61.0 \\\hline
    DHN'16 \cite{zhu2016deep} & CNN & \cite{cao2017hashnet} & 63.7 & 66.4 & 67.1 & 67.7 & 70.1 & 69.4 \\\hline
    HashNet'17 \cite{cao2017hashnet} & CNN & \cite{cao2017hashnet} & 66.2 & 69.9 & 71.6 & 68.7 & 71.8 & 73.6 \\\hline

    DeepBit'16 \cite{lin2016learning} & CNN & \cite{shen2020auto} & 39.2 & 40.3 & 42.9 & 40.7 & 41.9 & 43.0 \\ \hline
    BGAN'18 \cite{song2018binary} & GAN & \cite{shen2020auto} & 68.4 & 71.4 & 73.0 & 64.5 & 68.2 & 70.7 \\ \hline
    GreedyHash'18 \cite{su2018greedy} & CNN & \cite{shen2020auto} & 63.3 & 69.1 & 73.1 & 58.2 & 66.8 & 71.0 \\ \hline
    BinGAN'18 \cite{zieba2018bingan} & GAN & \cite{shen2020auto} & 65.4 & 70.9 & 71.3 & 65.1 & 67.3 & 69.6 \\ \hline
    DVB'19 \cite{shen2019unsupervised} & DVN & \cite{shen2020auto} & 60.4 & 63.2 & 66.5 & 57.0 & 62.9 & 62.3 \\ \hline
    DistillHash'19 \cite{yang2019distillhash} & SN & \cite{shen2020auto} & 66.7 & 67.5 & 67.7 & - & - & - \\ \hline
    TBH'20 \cite{shen2020auto} & AE & \cite{shen2020auto} & 71.7 & 72.5 & 73.5 & 70.6 & 73.5 & 72.2 \\ \hline

    CNNH'14 \cite{xia2014supervised} & CNN & \cite{xu2019dha} & 57.0 & 58.3 & 60.0 & 56.4 & 57.4 & 56.7 \\ \hline
    DNNH'15 \cite{lai2015simultaneous} & DNN & \cite{xu2019dha} & 59.8 & 61.6 & 63.9 & 59.3 & 60.3 & 61.0 \\ \hline
    DHN'16 \cite{zhu2016deep} & CNN & \cite{xu2019dha} & 63.7 & 66.4 & 67.1 & 67.7 & 70.1 & 69.4 \\ \hline
    HashNet'17 \cite{cao2017hashnet} & CNN & \cite{xu2019dha} & 66.3 & 69.9 & 71.6 & 68.7 & 71.8 & 73.6 \\ \hline
    DHA'19 \cite{xu2019dha} & CNN & \cite{xu2019dha} & 66.9 & 70.6 & 72.7 & 70.8 & 73.1 & 75.2 \\ \hline
    
    HashGAN'18 \cite{ghasedi2018unsupervised} & GAN & \cite{ghasedi2018unsupervised} & 71.5 & 73.7 & 74.8 & 69.7 & 72.5 & 74.4 \\ \hline
    UH-BDNN'16 \cite{do2016learning} & DNN & \cite{gu2019unsupervised} & 59.2 & 59.0 & 61.0 & - & - & -  \\ \hline
    UTH'17 \cite{huang2017unsupervised} & TN & \cite{gu2019unsupervised} & 54.3 & 53.7 & 54.7 & - & - & -  \\ \hline
    UDTH'19 \cite{gu2019unsupervised} & TN & \cite{gu2019unsupervised} & 64.4 & 67.7 & 69.6 & - & - & -  \\ \hline
    
    SSDH'17 \cite{yang2017supervised} & CNN & \cite{wang2020deep} & - & - & - & 69.7 & 72.5 & 74.4 \\\hline
    DPAH'20 \cite{wang2020deep} & PTN & \cite{wang2020deep} & - & - & - & 73.3 & 76.8 & \textbf{78.2} \\\hline

    DRDH'20 \cite{yang2020deep} & DQN & \cite{yang2020deep} & 80.5 & 81.7 & \textbf{81.8} & 71.5 & 74.8 & 76.1  \\\hline

    DVSQ'17 \cite{cao2017deep} & CNN & \cite{chen2019similarity} & 79.0 & 79.7 & - & 71.2 & 72.0 & -  \\\hline
    DTQ'18 \cite{liu2018deep} & TN & \cite{chen2019similarity} & 79.8 & 80.1 & - & 76.0 & 76.7 & - \\\hline
    SPDAQ'19 \cite{chen2019similarity} & CNN &  \cite{chen2019similarity} & \textbf{84.2} & 85.1 & - & \textbf{84.4} & \textbf{84.7} & -  \\\hline
    DSQ'19 \cite{eghbali2019deep} & CNN & \cite{eghbali2019deep} & 77.9 & 79.0 & 79.9 & - & - & -  \\\hline

     &&& \multicolumn{6}{c|}{CIFAR-10 Dataset} \\
    \hline
     & & & - & 12 Bits & 24 Bits & 32 Bits & 48 Bits & -\\
    \hline
    SDH'15 \cite{shen2015supervised} & PTN & \cite{zhai2020deep} &-& 45.4 & 63.3 & 65.1 & 66.0&-\\\hline
    DSH'16 \cite{liu2016deep} & CNN & \cite{zhai2020deep} &-& 64.4 & 74.2 & 77.0 & 79.9&-\\\hline
    DHN'16 \cite{zhu2016deep} & CNN & \cite{zhai2020deep} &-& 68.1 & 72.1 & 72.3 & 73.3 &-\\\hline
    DPSH'16 \cite{li2016feature} & SN & \cite{zhai2020deep} &-& 68.2 & 72.0 & 73.4 & 74.6 &- \\\hline
    DQN'16 \cite{cao2016deep} & CNN & \cite{zhai2020deep} &-& 55.4 & 55.8 & 56.4 & 58.0 &-\\\hline
    DSDH'17 \cite{li2017deep} & CNN & \cite{zhai2020deep} &-& 74.0 & 78.6 & 80.1 & 82.0 &- \\\hline
    ADSH'18 \cite{jiang2018asymmetric} & CNN & \cite{zhai2020deep} &-& 89.0 & 92.8 & 93.1 & 93.9 &-\\\hline
    DIHN2+ADSH'19 \cite{wu2019deep} & CNN & \cite{zhai2020deep} &-& 89.8 & 92.9 & 92.9 & 93.9 &-\\\hline
    DTH'20 \cite{zhai2020deep} & CNN & \cite{zhai2020deep} &-& \textbf{92.1} & \textbf{93.3} & \textbf{93.7} & \textbf{94.9} &-\\\hline
    \end{tabular}
    \label{tab:results5000}
\end{table}

\subsection{Applications}
The deep learning based approaches have been utilized for image retrieval pertaining to different applications such as cloth retrieval \cite{lin2015rapid}, 
biomedical image retrieval 
\cite{dubey2019local}, face retrieval \cite{dong2018deep}, \cite{dubey2018average}, 
remote sensing image retrieval \cite{liu2020similarity}, landmark retrieval \cite{yang2019dame}, 
social image retrieval \cite{zhu2020dual}, and video retrieval \cite{zhang2016play}.

\begin{table}[!t]
    \centering
    \caption{Mean Average Precision (mAP) with 1000 retrieved images (mAP@1000) in \% for different deep learning based image retrieval methods over ImageNet, CIFAR-10 and MNIST datasets.}

    \begin{tabular}{|m{0.128\textwidth}|m{0.04\textwidth}|m{0.03\textwidth}|m{0.03\textwidth}|m{0.03\textwidth}|m{0.03\textwidth}|m{0.03\textwidth}|}
    \hline
    Method & Network Type & Result Source & 16 Bits & 32 Bits & 48 Bits & 64 Bits\\ 
    \hline
    &&&\multicolumn{4}{|c|}{ImageNet Dataset}\\
    \hline
    CNNH'14 \cite{xia2014supervised} & CNN & \cite{cao2017hashnet} & 28.1 & 45.0 & 52.5 & 55.4 \\\hline
    SDH'15 \cite{shen2015supervised} & PTN & \cite{cao2017hashnet} & 29.9 & 45.5 & 55.5 & 58.5 \\\hline
    DNNH'15 \cite{lai2015simultaneous} & DNN & \cite{cao2017hashnet} & 29.0 & 46.1 & 53.0 & 56.5 \\\hline
    DHN'16 \cite{zhu2016deep} & CNN & \cite{cao2017hashnet} & 31.1 & 47.2 & 54.2 & 57.3 \\\hline
    HashNet'17 \cite{cao2017hashnet} & CNN & \cite{cao2017hashnet} & 50.6 & 63.1 & 66.3 & 68.4 \\\hline
    
    SSDH'17 \cite{yang2017supervised} & CNN & \cite{wang2020deep} & 63.4 & 69.2 & 70.1 & 70.7 \\\hline
    DSQ'19 \cite{eghbali2019deep} & CNN & \cite{eghbali2019deep} & 57.8 & 65.4 & 68.0 & 69.4 \\ \hline
    DPAH'20 \cite{wang2020deep} & PTN & \cite{wang2020deep} & \textbf{65.2} & \textbf{70.0} & \textbf{71.5} & \textbf{71.4} \\\hline
    
    &&&\multicolumn{4}{|c|}{CIFAR-10 Dataset}\\
    \hline    
    BGAN'18 \cite{song2018binary} & GAN & \cite{shen2020auto} & 52.5 & 53.1 & - & 56.2 \\ \hline
    GreedyHash'18 \cite{su2018greedy} & CNN & \cite{shen2020auto} & 44.8 & 47.3 & - & 50.1 \\ \hline
    BinGAN'18 \cite{zieba2018bingan} & GAN & \cite{shen2020auto} & 47.6 & 51.2 & - & 52.0 \\ \hline
    HashGAN'18 \cite{ghasedi2018unsupervised} & GAN & \cite{shen2020auto} & 44.7 & 46.3 & - & 48.1 \\ \hline
    DVB'19 \cite{shen2019unsupervised} & DVN & \cite{shen2020auto} & 40.3 & 42.2 & - & 44.6 \\ \hline
    DistillHash'19 \cite{yang2019distillhash} & SN & \cite{shen2020auto} & 28.4 & 28.5 & - & 28.8 \\ \hline
    TBH'20 \cite{shen2020auto} & AE & \cite{shen2020auto} & \textbf{53.2} & \textbf{57.3} & - & \textbf{57.8} \\ \hline

    SDH'15 \cite{erin2015deep} & PTN & \cite{erin2015deep} & 18.8 & 20.8 & - & 22.5 \\ \hline
    
    DAR'16 \cite{huang2016unsupervised} & TN & \cite{huang2016unsupervised} & 16.8 & 17.0 & - & 17.2 \\ \hline
    
    DH'15 \cite{erin2015deep} & DNN & \cite{huang2017unsupervised} & 16.2 & 16.6 & - & 17.0 \\ \hline
    DeepBit'16 \cite{lin2016learning} & CNN & \cite{huang2017unsupervised} & 19.4 & 24.9 & - & 27.7 \\ \hline
    UTH'17 \cite{huang2017unsupervised} & TN & \cite{huang2017unsupervised} & 28.7 & 30.7 & - & 32.4 \\ \hline
    DBD-MQ'17 \cite{duan2017learning} & CNN & \cite{duan2017learning} & 21.5 & 26.5 & - & 31.9 \\ \hline
    UCBD'18 \cite{lin2018unsupervised} & CNN & \cite{lin2018unsupervised} & 26.4 & 27.9 & - & 34.1\\ \hline
    UH-BDNN'16 \cite{do2016learning} & DNN & \cite{gu2019unsupervised} & 30.1 & 30.9 & - & 31.2 \\ \hline
    UDTH'19 \cite{gu2019unsupervised} & TN & \cite{gu2019unsupervised} & 46.1 & 50.4 & - & 54.3 \\ \hline
    
    &&&\multicolumn{4}{|c|}{MNIST Dataset} \\
    \hline
    SDH'15 \cite{erin2015deep} & PTN & \cite{erin2015deep} & \textbf{46.8} & \textbf{51.0} & - & \textbf{52.5} \\ \hline
    DH'15 \cite{erin2015deep} & DNN & \cite{huang2017unsupervised} & 43.1 & 45.0 & - & 46.7 \\ \hline
    DeepBit'16 \cite{lin2016learning} & CNN & \cite{huang2017unsupervised} & 28.2 & 32.0 & - & 44.5 \\ \hline
    UTH'17 \cite{huang2017unsupervised} & TN & \cite{huang2017unsupervised} & 43.2 & 46.6 & - & 49.9 \\ \hline
    
    \end{tabular}
    \label{tab:results1000}
\end{table}

\subsection{Others}
The hashing difficulty is also increased by generating the harder samples in a self-paced manner \cite{jin2019ssah} to make the network training as reasoning oriented.
In the initial work, the pre-trained CNN features have also very promising retrieval performance \cite{babenko2014neural}. 
Recently, the transfer learning has been also utilized in \cite{zhai2020deep} for deep transfer hashing.

\subsection{Summary}
Researchers have come up with various loss functions to facilitate the discriminative learning of features by the networks for image retrieval. The losses constraint and guide the training of the deep learning models.
    The image retrieval has shown a great utilization with its application to solve the real-life problems. Researchers have also tried to understand what works and what not for deep learning based image retrieval. The transfer learning has been also utilized for retrieval.

\begin{table}[!t]
    \centering
    \caption{mAP@54000 and mAP@All in \% for state-of-the-art and recent image retrieval methods over the CIFAR-10 dataset.  }
    \begin{tabular}{|m{0.124\textwidth}|m{0.04\textwidth}|m{0.03\textwidth}|m{0.02\textwidth}|m{0.02\textwidth}|m{0.02\textwidth}|m{0.02\textwidth}|m{0.02\textwidth}|}
    \hline
     &  &  & \multicolumn{5}{c|}{CIFAR-10 Dataset}\\
    \cline{4-8}
    Method Name & Network Type & Result Source & 16 Bits & 24 Bits & 32 Bits & 48 Bits & 64 Bits \\ 
    \hline
    &&& \multicolumn{5}{|c|}{mAP@54000} \\\hline
    CNNH'14 \cite{xia2014supervised} & CNN & \cite{xu2019dha} & 47.6 &-& 47.2 & 48.9 & 50.1 \\ \hline
    DNNH'15 \cite{lai2015simultaneous} & TN & \cite{xu2019dha} & 55.9 &-& 55.8 & 58.1 & 58.3 \\ \hline
    SDH'15 \cite{shen2015supervised} & PTN & \cite{xu2019dha} & 46.1 &-& 52.0 & 55.3 & 56.8 \\ \hline
    DHN'16 \cite{zhu2016deep} & CNN & \cite{xu2019dha} & 56.8 && 60.3 & 62.1 & 63.5 \\ \hline
    HashNet'17 \cite{cao2017hashnet} & CNN & \cite{xu2019dha} & 64.3 &-& 66.7 & 67.5 & 68.7 \\ \hline
    DHA'14 \cite{xu2019dha} & CNN & \cite{xu2019dha} & 65.2 && 68.1 & 69.0 & 69.9 \\\hline
    HashGAN'18 \cite{ghasedi2018unsupervised} & GAN & \cite{ghasedi2018unsupervised} & 66.8 &-& 73.1 & 73.5 & 74.9 \\ \hline
    DTQ'18 \cite{liu2018deep} & TN & \cite{liu2018deep} & \textbf{78.9} &-& 79.2 & - & - \\ \hline
    DRDH'20 \cite{yang2020deep} & DQN & \cite{yang2020deep} & 78.7 &-& \textbf{80.5} & \textbf{80.6} & \textbf{80.3} \\ \hline
    
    &&& \multicolumn{5}{|c|}{mAP@All} \\\hline

    DQN'16 \cite{cao2016deep} & CNN & \cite{chen2019similarity} & - & 55.8 & 56.4 & 58.0 & - \\\hline
    DPSH'16 \cite{li2016feature} & SN & \cite{chen2019similarity} & - & 72.7 & 74.4 & 75.7 & - \\\hline
    DSDH'17 \cite{li2017deep} & CNN & \cite{chen2019similarity} & - & 78.6 & 80.1 & 82.0 & - \\\hline
    DTQ'18 \cite{liu2018deep} & DQN & \cite{chen2019similarity} & - & 79.0 & 79.2 & - & - \\\hline
    DVSQ'17 \cite{cao2017deep} & CNN & \cite{chen2019similarity} & - & 80.3 & 80.8 & 81.1 & - \\\hline
    SPDAQ'19 \cite{chen2019similarity} & CNN & \cite{chen2019similarity} & - & \textbf{88.4} & \textbf{89.1} & \textbf{89.3} & - \\\hline
    SSAH'19 \cite{jin2019ssah} & GAN & \cite{jin2019ssah} & - & 87.8 & - & 88.6 & - \\\hline
    DeepBit'19 \cite{lin2016learning} & CNN & \cite{deng2019unsupervised} & 22.0 & - & 24.1 & - & 29.0 \\ \hline
    BGAN'19 \cite{song2018binary} & GAN & \cite{deng2019unsupervised} & 49.7 & - & 47.0 & - & 50.7 \\ \hline
    UADH'19 \cite{deng2019unsupervised} & GAN & \cite{deng2019unsupervised} & \textbf{67.7} & - & 68.9 & - & \textbf{69.6} \\ \hline
    DSAH'19 \cite{ge2019deep} & CNN & \cite{ge2019deep} & - & 84.1 & 84.5 & 84.9 & - \\ \hline
    
    \end{tabular}
    \label{tab:results_all}
\end{table}

\section{Performance Comparison}
\label{performance_comparison}
This survey also presents a performance analysis for the state-of-the-art deep learning based image retrieval approaches. The Mean Average Precision (mAP) reported for the different image retrieval approaches is summarized in Table \ref{tab:results5000}, \ref{tab:results1000}, and \ref{tab:results_all}. The mAP@5000 (i.e., 5000 retrieved images) using various existing deep learning approaches is summarized in Table \ref{tab:results5000} over CIFAR-10, NUS-WIDE and MS COCO datasets. The results over CIFAR-10, ImageNet and MNIST datasets using different state-of-the-art deep learning based image retrieval methods are compiled in Table \ref{tab:results1000} in terms of the mAP@1000. The mAP@54000 using few methods is reported in Table \ref{tab:results_all} over the CIFAR-10 dataset. The standard mAP is also depicted in Table \ref{tab:results_all} by considering all the retrieved images for CIFAR-10 dataset using some of the available literature. Note that 2nd column in Table \ref{tab:results5000}, \ref{tab:results1000}, and \ref{tab:results_all} list the source reference of the corresponding method reported results. Following are the observations out of these results by deep learning methods:
\begin{itemize}
    \item Recently proposed Deep Transfer Hashing (DTH) by Zhai et al. (2020) \cite{zhai2020deep} have shown outstanding performance over CIFAR-10 and NUS-WIDE datasets in terms of the mAP@5000. Other promising methods include Deep Spatial Attention Hashing (DSAH) by Ge et al. (2019) \cite{ge2019deep}, Similarity Preserving Deep Asymmetric Quantization (SPDAQ) by Chen et al. (2019) \cite{chen2019similarity}, Deep  Position-Aware Hashing (DPAH) by Wang et al. (2020) \cite{wang2020deep} and Deep Reinforcement De-Redundancy Hashing (DRDH) by Yang et al. (2020) \cite{yang2020deep}.
    
    \item The Twin-Bottleneck Hashing (TBH) introduced by Shen et al. (2020) \cite{shen2020auto} is also observed as an appealing method using autoencoder having a double bottleneck over the CIFAR-10 dataset in terms of the mAP@1000. However, the Deep Position-Aware Hashing (DPAH) investigated by Wang et al. (2020) \cite{wang2020deep} have outperformed the other approaches over ImageNet dataset.
    Supervised Deep Hashing (SDH) by Erin et al. (2015) \cite{erin2015deep} has depicted appealing performance over the MNIST dataset.
    
    \item The deep reinforcement learning based image retrieval model, namely Deep Reinforcement De-Redundancy Hashing (DRDH) by Yang et al. (2020) \cite{yang2020deep}, is one of recent breakthrough as supported by superlative mAP@54000 over the CIFAR-10 dataset. The Deep Triplet Quantization \cite{liu2018deep} is also one of the favourable model for feature learning.
    
    \item The Similarity  Preserving  Deep  Asymmetric Quantization (SPDAQ) by Chen et al. (2019) \cite{chen2019similarity} and Unsupervised ADversarial Hashing (UADH) by Deng et al. (2019) \cite{deng2019unsupervised} methods have been also identified as very encouraging based on the mAP by considering all the retrieved images over the CIFAR-10 dataset.
\end{itemize}

\section{Conclusion and Future Directives}
\label{conclusion}
\subsection{Conclusion and Trend}
This paper presents a comprehensive survey of deep learning methods for content based image retrieval. As most of the deep learning based developments are recent, this survey majorly focuses over the image retrieval methods using deep learning in a decade from 2011 to 2020. A detailed taxonomy is presented in terms of different supervision type, different networks used, different data type of descriptors, different retrieval type and other aspects. The detailed discussion under each section is also presented with the further categorization. A chronological summarization is presented to show the evolution of the deep learning models for image retrieval. Moreover, the chronological overview is also portrayed under each category to showcase the growth of image retrieval approaches. A summary of large-scale common datasets used for image retrieval is also compiled in this survey. A performance analysis of the state-of-the-art deep learning based image retrieval methods is also conducted in terms of the mean average precision for different no. of retrieved images.

The research trend in image retrieval suggests that the deep learning based models are driving the progress. The recently developed models such as generative adversarial networks, autoencoder networks and reinforcement learning networks have shown the superior performance for image retrieval. The discovery of better objective functions has been also the trend in order to constrain the learning of the hash code for discriminative, robust and efficient image retrieval. The semantic preserving class-specific feature learning using different networks and different quantization techniques is also the recent trend for image retrieval. Other trends include utilization of attention module, transfer learning, etc.

\subsection{Future Directions}
The future work in image retrieval using deep learning can include exploration of improved deep learning models, more relevant objective functions, minimum loss based quantization techniques, semantic preserving feature learning, and attention focused feature learning.
The future direction in the image retrieval might be driven from the basic goal of the expected solution. There are three important aspects of any retrieval system, which include the discriminative ability, robustness capability and fast image search. In order to achieve the discriminative ability, the features corresponding to the samples of different class should be as far as possible. Thus, different approaches such as triplet based objective function, consideration of class distribution, incorporation of distance between class centroids, etc. can be exploited. In order to maintain the robustness property, various data augmentation, layer manipulation, siamese loss based objective functions, feature normalization, incorporation of class distribution and majority voting in the feature representation, etc. can be explored. In order to perform the faster image search, the learnt feature or hash code should be as low dimensional and compact as possible. Thus, better strategy for feature quantization and maximizing the relevant information into feature space in a compact way can be seen as one of the future directions. The self-supervised learning has shown very promising performance for different down-stream tasks and has potential to learn the important features in compact form. Thus, in future, the self-supervised learning can boost the performance of image retrieval models significantly.

\section*{Acknowledgement}
This work is supported by Global Innovation \& Technology Alliance (GITA) on behalf of Department of Science and Technology (DST), Govt. of India through project no. GITA/DST/TWN/P-83/2019.

{\small
\bibliographystyle{IEEEtran}
\bibliography{Ref}
}

\begin{IEEEbiography}[{\includegraphics[width=1in,height=1.25in,clip,keepaspectratio]{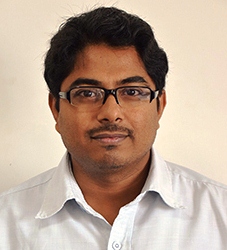}}]{Shiv Ram Dubey}
has been with the Indian Institute of Information Technology (IIIT), Sri City since June 2016, where he is currently the Assistant Professor of Computer Science and Engineering. He received the Ph.D. degree in Computer Vision and Image Processing from Indian Institute of Information Technology, Allahabad (IIIT Allahabad) in 2016. Before that, from August 2012-Feb 2013, he was a Project Officer in the Computer Science and Engineering Department at Indian Institute of Technology, Madras (IIT Madras). 
He was a recipient of several awards, including the Best PhD Award in PhD Symposium, IEEE-CICT2017 at IIITM Gwalior and NVIDIA GPU Grant Award Twice from NVIDIA. 
His research interest includes Computer Vision, Deep Learning, Image Feature Description, and Content Based Image Retrieval.  
\end{IEEEbiography}

\end{document}